\pdfoutput=1

\documentclass[11pt]{article}

\usepackage[preprint]{acl}

\usepackage{times}
\usepackage{latexsym}

\usepackage[T1]{fontenc}

\usepackage[utf8]{inputenc}

\usepackage{microtype}

\usepackage{inconsolata}

\usepackage{graphicx}

\usepackage{enumitem}
\usepackage{algorithm}
\usepackage{algorithmic}

\usepackage{amsmath}
\usepackage{amssymb}
\usepackage{mathtools}
\usepackage{amsthm}
\usepackage[capitalize,noabbrev]{cleveref}

\title{LLMs learn governing principles of dynamical systems, revealing an in-context neural scaling law}

\author{
  \textbf{Toni J.B. Liu}\textsuperscript{$*$},\,
  \textbf{Nicolas Boullé}\textsuperscript{$\dagger$},\,
  \textbf{Raphaël Sarfati}\textsuperscript{$*$},\,
  \textbf{Christopher J. Earls}\textsuperscript{$*$}
\\
\\
  \textsuperscript{$*$}Cornell University, USA,\,
  \textsuperscript{$\dagger$}Imperial College London, UK
\\
  \small{
    \textbf{Correspondence:} \href{mailto:jl3499@cornell.edu}{jl3499@cornell.edu}
  }
}

\begin{document}
\maketitle
\begin{abstract}
  We study LLMs' ability to extrapolate the behavior of various dynamical systems, 
  including stochastic, chaotic, continuous, and discrete systems,
  whose evolution is governed by principles of physical interest.
  Our results show that LLaMA-2, a language model trained on text, achieves accurate predictions of dynamical system time series without fine-tuning or prompt engineering. 
  Moreover, the accuracy of the learned physical rules increases with the length of the input context window, revealing an in-context version of a neural scaling law. 
  Along the way, we present a flexible and efficient algorithm for extracting probability density functions of multi-digit numbers directly from LLMs.\footnote{Our code is publicly available at \url{https://github.com/AntonioLiu97/llmICL}.}
\end{abstract}



\section{Introduction}
\label{sec_intro}


Since the introduction of the transformer architecture~\cite{vaswani2017attention}, Large language models (LLMs) have shown a variety of unexpected emergent properties, such as program execution~\cite{nye2021show} and multi-step reasoning~\cite{suzgun2022challenging}.

\begin{figure}[htbp]
  \begin{center}
    \centerline{\includegraphics[width=\columnwidth]{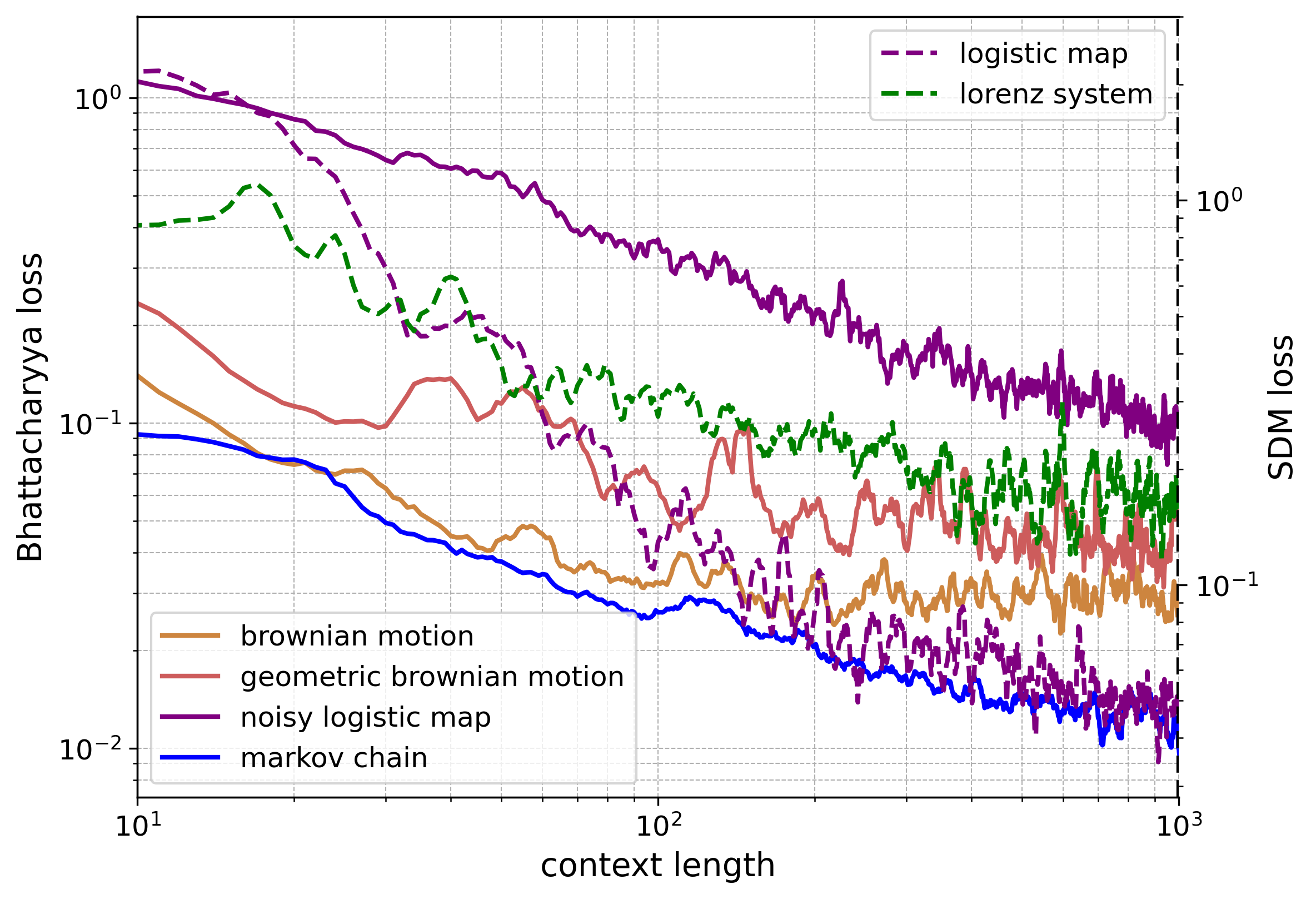}}
    \caption{Evolution of the loss function for the predicted next state by LLaMA-13b with respect to the number of observed states in various physical systems. We employ the Bhattacharyya distance as a loss function for stochastic systems (solid lines), and the squared deviations from the mean (SDM) for deterministic systems (dashed lines).
      Brownian motion and geometric Brownian motion deviate significantly from power law scaling, which can be explained by their lack of stationary distributions
      (\cref{app: invariant measure}).
    }
    \label{fig:ICL}
    \vspace{-2em}
  \end{center}
\end{figure}

Our work explores LLMs' ability to model the world by proposing a new perspective and empirical approach. Inspired by the recent observations that LLMs are capable of in-context time series extrapolation without specific prompting or fine-tuning~\cite{gruver2023large,jin2023time}, we aim to quantify LLMs' ability to extrapolate stochastic dynamical systems.
We find that, as the number of observed time steps increases, an LLM's statistical prediction consistently converges to the ground truth transition rules underlying the system; leading to an empirical scaling law, as observed in~\cref{fig:ICL}.

\noindent Our \textbf{main contributions} are as follows:
\begin{itemize}[noitemsep,topsep=-\parskip,leftmargin=*]
  \item demonstrating LLMs' zero-shot ability to model the evolution of dynamical systems without instruction prompting;
  \item implementing a computationally efficient framework called \emph{Hierarchy-PDF} to extract statistical information of a dynamical system learned by a transformer-based LLM;
  \item discovering a scaling law between the accuracy in the learned transition rules (compared to ground truth) and the context window length.
\end{itemize}

\section{Background and related work}
\label{sec:background}

\emph{In-context learning} refers to LLM's emergent ability to learn from examples included in the prompt~\cite{brown2020language}. One example of in-context learning is zero-shot time series forecasting~\cite{gruver2023large}. This work aims to forecast empirical time series and introduces a tokenization procedure to convert a sequence of floating point numbers into appropriate textual prompts for LLMs. This led to several subsequent studies on the application of LLMs for time series forecasting~\cite{chen2023gatgpt,jin2023large,jin2023time,dooley2023forecastpfn,schoenegger2023large,wang2023prompt,xu2023transformer}.

Unlike these prior studies, our work does not focus on forecasting real-world time series, such as weather data or electricity demand, where the underlying model generating the sequence is unavailable or undefined.
Instead, we aim to \emph{extract the learned transition rules from the probability vector generated by the LLM} and compare them against the ground truth rules (chaotic, stochastic, discrete, continuous, etc.) governing the input time series.

\section{Methodology}
\label{sec:methods}
Our methodology for testing LLMs' ability to learn physical rules from in-context data consists of three steps:
\begin{enumerate}[noitemsep,topsep=-\parskip,leftmargin=*]
  \item Sample a time series $\{x_t\}_{t\geq 0}$ from a given dynamical system governed by Markovian transition rules $P_{ij}$.
  \item Prompt the LLM with this time series to extract the learned probability densities for subsequent digits $\tilde{P}_{ij}$.
  \item Measure the discrepancy, between the ground truth $P_{ij}$ and learned $\tilde{P}_{ij}$, using Bhattacharyya distance.\footnote{Other loss functions may be appropriate depending on whether the dynamical system is stochastic or deterministic, see \cref{loss_functions}}
\end{enumerate}
\vskip \parskip

\subsection{Prompt generation}

\paragraph{Markov processes.}
Most of our testing data may be modeled as discrete-time Markov chains, where the probability density function (PDF) of the next state at time $t+1$ depends solely on the previous state, $x_t$, at time $t$:
\[ P(X_{t+1} | x_1, \dots, x_t) = P(X_{t+1} | x_t).\]
This models either discrete iterative systems or continuous dynamical systems after time-discretization.

\paragraph{Time series tokenization.}
An input time series typically consists of ($\sim 10^3$) time steps, each represented as a real number. We first rescale each number and represent it using $n$ digits (typically, $n=3$). Each time series is rescaled to the interval $[1.50, 8.50]$ so that the number of digits never changes throughout the series.
We then follow the scheme introduced in~\cite{gruver2023large} to serialize the time series as strings and tokenize them.

\subsection{Extraction of transition rules} \label{sec: extract transition rules}

\paragraph{Discrete state space.} \label{method:discrete}
When the Markov process is discrete and has a finite state space, each state can be represented by a single token.
We employ tokens corresponding to the ten number strings: $0,\ldots,9$.
We find that even the most sophisticated LLaMA model (LLaMA-70b) can only learn up to 9 discrete states. Therefore, we do not attempt to go beyond 9 distinct states by extending to non-number tokens (see~\cref{Markov chains with larger numbers of states} for more details).

\begin{figure}[htbp]
  \begin{center}
    \centerline{\includegraphics[width=\columnwidth]{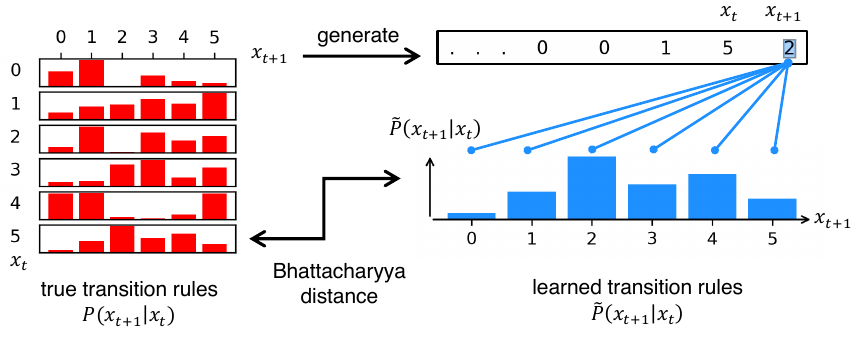}}
    \caption{Extracting learned transition rules of systems with discrete state space.}
    \label{fig:discrete}
  \end{center}
  \vskip -0.2in
\end{figure}

\cref{fig:discrete} illustrates our framework for learning discrete Markov chains with LLMs. First, we randomly sample an $n \times n$ transition matrix $(P_{ij})$. We then generate a Markov chain according to $P_{ij}$, tokenize the time series and pass to an LLM with no additional ``prompt engineering". The length of the series is chosen such that the tokenized representation does not exceed the length of the LLM's context window. We extract the LLM's prediction for the next state by performing a softmax operation on the output logits corresponding to the $n$ allowed states and discarding all other logits.

\paragraph{Continuous state space.}
Many stochastic processes, such as the Brownian motion~\cite{einstein1905molekularkinetischen,perrin1909mouvement}, are supported on continuous state space.
For these processes, we represent the value of each state as a multi-digit number and separate each state using the comma symbol ``,''.
As observed in~\cite{jin2023time}, an LLM prediction of multi-digit values can be naturally interpreted as a hierarchical softmax distribution~\cite{mnih2008scalable,challu2023nhits}.

\begin{figure}[htbp]
  \begin{center}
  \vspace{-1em}
    \centerline{\includegraphics[width=\columnwidth]{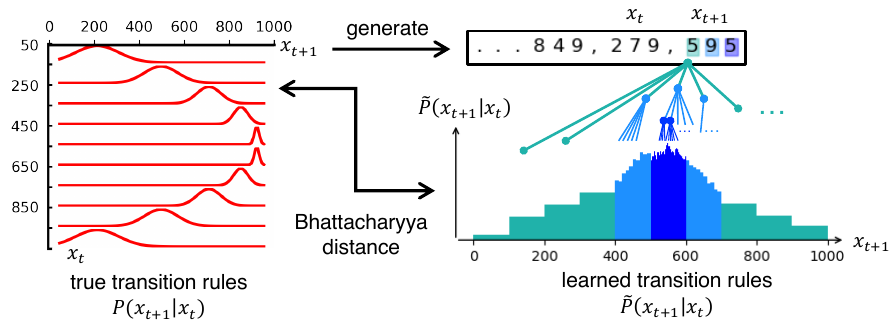}}
    \caption{An example of hierarchical transition rules extracted from LLaMa-13b.
      The PDF bins are color-coded based on resolutions, which in this example are more refined near the mode.
      The height of $\tilde{P}(x_{t+1}|x_t)$ is shown in log scale.}
    \label{fig:continuous}
    \vspace{-2.5em}
  \end{center}
\end{figure}

Specifically, let $u$ denote a multi-digit string representing the value of a state at a given time-step, then the LLM's softmax prediction for the $i^{th}$ digit, $u_{i}$, provides a histogram of ten bins of width $0.1^{i}$. Subsequently, the prediction of the $(i+1)^\mathrm{th}$ digit goes down one level into the hierarchical tree by refining one of the bins into ten finer bins of width $0.1^{i+1}$, and so on until the last digit is processed (see~\cref{fig:continuous}). The top right of the figure shows an example of a time series serialized as an input string.

\paragraph{Hierarchy-PDF algorithm.}
While a single pass through the LLM yields a discretized PDF represented by bins of various widths, we can refine the PDF by querying each coarse bin. For example, to furnish a maximal resolution PDF of a 3-digit number, we need to query all $10^2$ combinations of the first two digits of that number. Suppose a time series consists of $S$ values (steps), each represented as $n$ digits. Obtaining a maximal resolution PDF for each value of the entire sequence requires $10^{n-1}S$ forward passes of the LLM. This daunting process could be significantly simplified because most of the $10^{n-1}S$ inputs differ only in the last tokens, and thus one can recursively cache the key and value matrices associated with the shared tokens. The computation can be further reduced by refining only the high-probability bins near the mode, which dominate the loss functions, as shown in \cref{fig:continuous}. \cref{recursive refiner simple} outlines the \emph{Hierarchy-PDF} algorithm used to recursively refine the PDF associated with a multi-digit value in a time series (more details available in~
\cref{Appendix}).

\begin{algorithm}[htbp]
  \caption{Hierarchy-PDF}
  \label{recursive refiner simple}
  \begin{algorithmic}
    \STATE {\bfseries Input:}  Unrefined PDF, current depth $D_c$, target depth $D_t$
    \STATE {{\bfseries Procedure:} RecursiveRefiner(PDF, $D_c$, $D_t$)}
    \begin{ALC@g}
      \IF{$D_c=D_t$}
      \STATE{end the recursion}
      \ELSIF{current branch is refined}
      \STATE Alter the last digit to launch 9 recursive branches
      \STATE{RecursiveRefiner(PDF, $D_c$, $D_t$)}
      \ELSIF{current branch is unrefined}
      \STATE{refine PDF with new logits}

      \IF{$D_c+1 < D_t$}
      \STATE Append the last digit to launch 10 recursive branches
      \STATE{RecursiveRefiner(PDF, $D_c+1$, $D_t$)}
      \ENDIF
      \ENDIF
    \end{ALC@g}
    \STATE {\bfseries Output:} Refined PDF
  \end{algorithmic}
\end{algorithm}

\section{Experiments and Results} \label{sec:experiments}
This section reports empirical in-context learning results on two example systems: a discrete Markov chain and a stochastic logistic map. We defer discussion of other systems, reported in \cref{fig:ICL}, to \cref{additional_experiments_stochastic,additional_experiments_deterministic}. Each experiments is repeated ten times with trajectories  initiated by different random seeds.

\paragraph{Model choice.}
All numerical experiments reported in this section are performed using the open-source LLaMA-13b model.
While we observe that larger language models, such as LLaMA-70b, may achieve lower in-context loss on some dynamical systems (\cref{Markov chains with larger numbers of states}), they do not display qualitative differences that would affect our conclusions.

\subsection{Markov chains with discrete states}
The transition rules of a time-independent Markov chain with $n$ states consist of a stochastic matrix $(P_{ij})_{1\leq i,j\leq n}$, defined as
\vspace{-0.5em}
\[
  P_{ij} = P(X_{t+1}=j|X_t=i), \quad 1\leq i,j\leq n.
\]

Using the testing procedures elaborated on in \cref{method:discrete}, we generate 10 Markov chains, each from a distinct and randomly generated transition matrix of size $n=4$.

\begin{figure}[htbp]
  \begin{center}    \centerline{\includegraphics[width=\columnwidth]{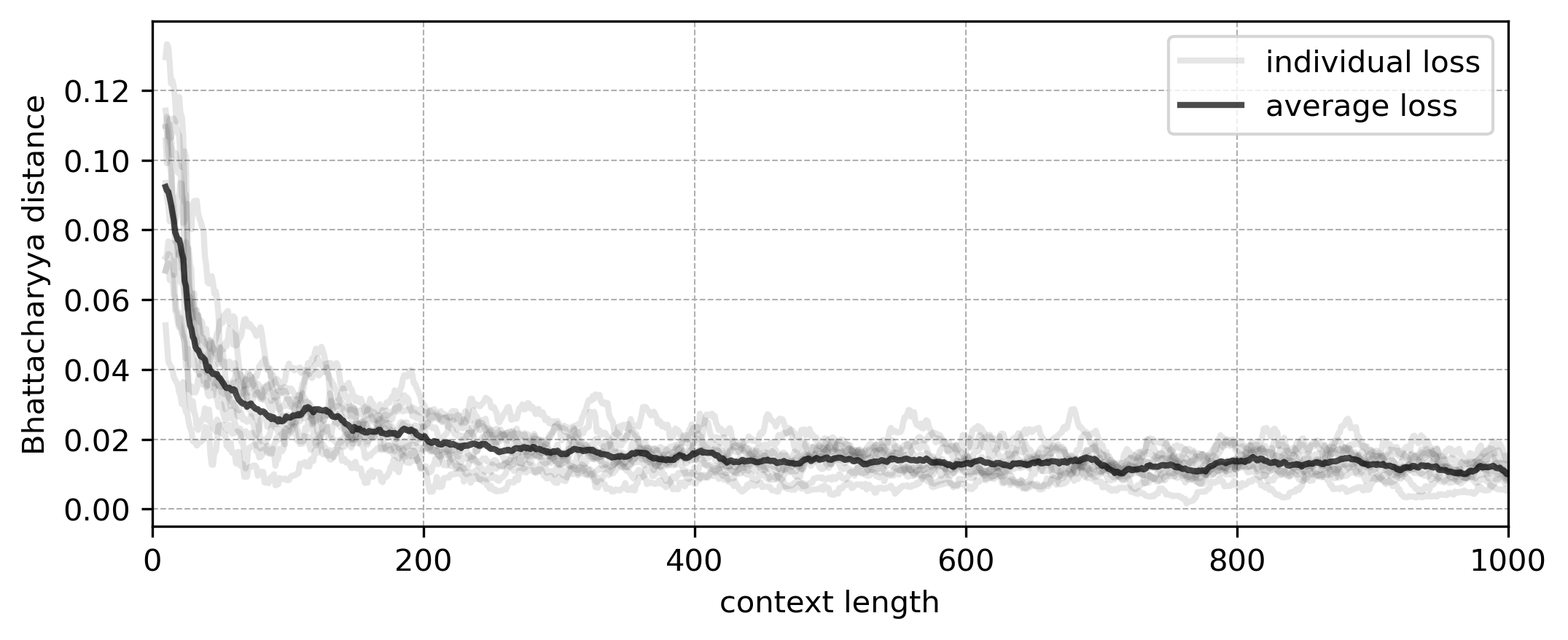}}
    \vspace{-1em}
    \caption{Markov chain in-context loss curves decay rapidly with respect to the input time series length. The average loss is obtained from 10 individual loss curves.}
    \label{fig:MC ICL}
  \end{center}
  \vspace{-2em}
\end{figure}

The corresponding loss curves between the LLM predictions and the ground truth are displayed in \cref{fig:MC ICL}.
The LLM formulates remarkably accurate statistical predictions as more time steps are observed in context, even though the transition rules are synthesized completely at random.
These conclusions hold for larger transition matrices ($n>4$) and more sophisticated LLMs, such as LLaMA-70b (see \cref{Markov chains with larger numbers of states}).

\subsection{Noisy logistic map}
The logistic map, first proposed as a discrete-time model for population growth, is one of the simplest dynamical systems that manifest chaotic behaviors~\cite{strogatz2018nonlinear}. It is governed by the following iterative equation:
\vspace{-0.5em}
\[
  x_{t+1} = f(x_t) = r x_t(1-x_t),\quad x_0 \in (0,1),
\]

where $r\in[1,4)$ is a parameter that controls the system's behavior.\footnote{In our experiments, we set $r=3.9$, which places the system firmly in the chaotic regime. 
At this value, the logistic map exhibits sensitive dependence on initial conditions and aperiodic behavior characteristic of chaos.}
The logistic map system becomes stochastic when one introduces small Gaussian perturbations of variance at each step, resulting in modified iterative equation:
\vspace{-0.5em}
\[
x_{t+1} = f(x_{t} + \epsilon), 
\]

where $\epsilon \stackrel{\text{i.i.d}}{\sim} \mathcal{N}(0, \sigma^2)$.
In this case, the ground truth distribution of the next state, $x_{t+1}$, 
conditioned on the current state $x_t$ is Gaussian with mean $f(x_t)$ and variance $(\sigma f'(x_t))^2$:
\begin{equation} \label{eq: noisy logistic transition function}
  X_{t+1}|\{X_t = x_{t}\} \sim \mathcal{N}\left(f(x_t), (\sigma f'(x_t))^2\right).
\end{equation}
The first derivative of $f$ measures how sensitive the local dynamics are to external perturbations. 
This intuitively explains why the standard deviation of the conditional distribution should be proportional to $f'$.\footnote{Note that the approximation in \cref{eq: noisy logistic transition function} assumes a small perturbation compared to the second derivative. That is, $\sigma^2 \ll 1/f''(x)$.}
We visualize the remarkable accuracy of the LLM's in-context predictions compared to the ground-truth transition rules in Figure \ref{fig:noisy_logistic_example_snapshot_988} as well as Figure \ref{noisy_logistic_map_snapshots} in the appendix.

\begin{figure}[t]
  \begin{center}
    \centerline{\includegraphics[width=\columnwidth]{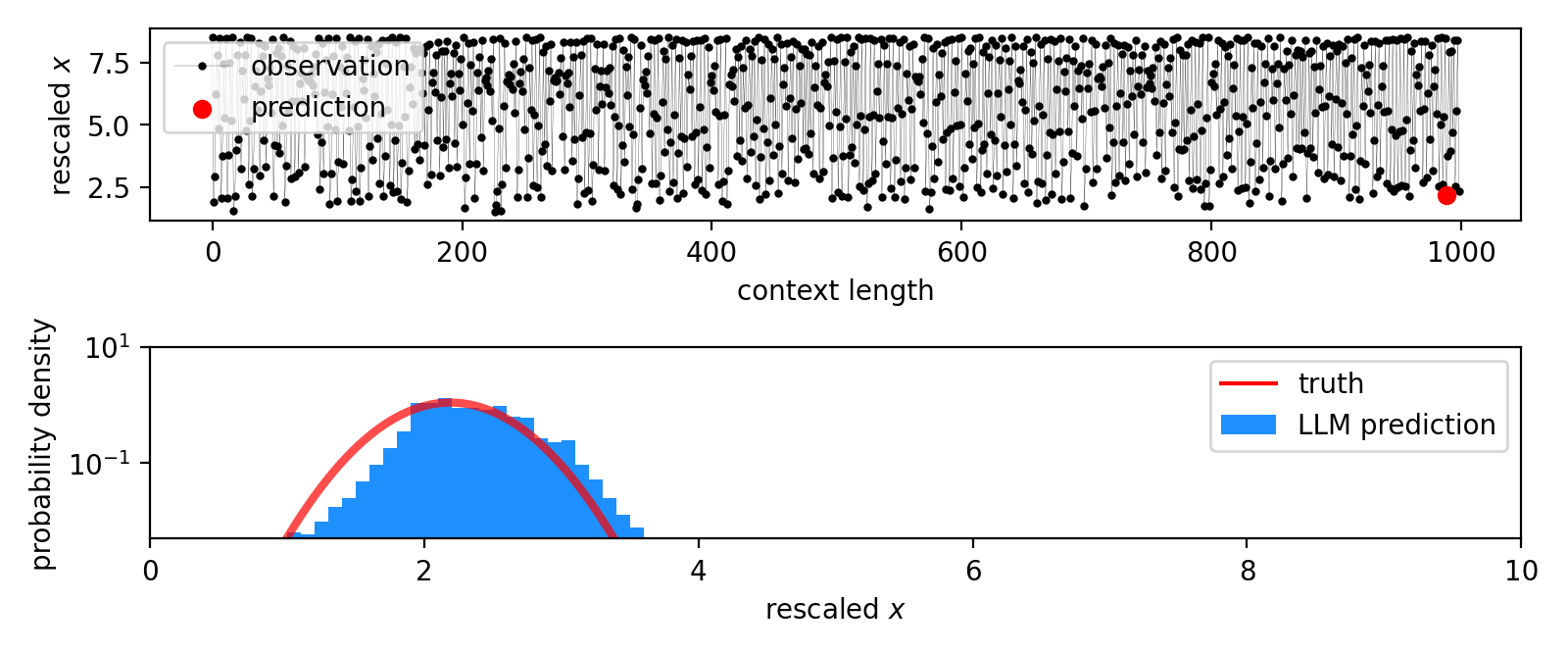}}
    \vspace{-1em}
    \caption{LLM's prediction for the $980^{th}$ state of a noisy logistic map,
    conditioned on observing the first $979$ states in-context, 
    compared with the ground-truth transition probability function.}
    \label{fig:noisy_logistic_example_snapshot_988}
  \end{center}
  \vspace{-1em}

  \begin{center}
    \centerline{\includegraphics[width=\columnwidth]{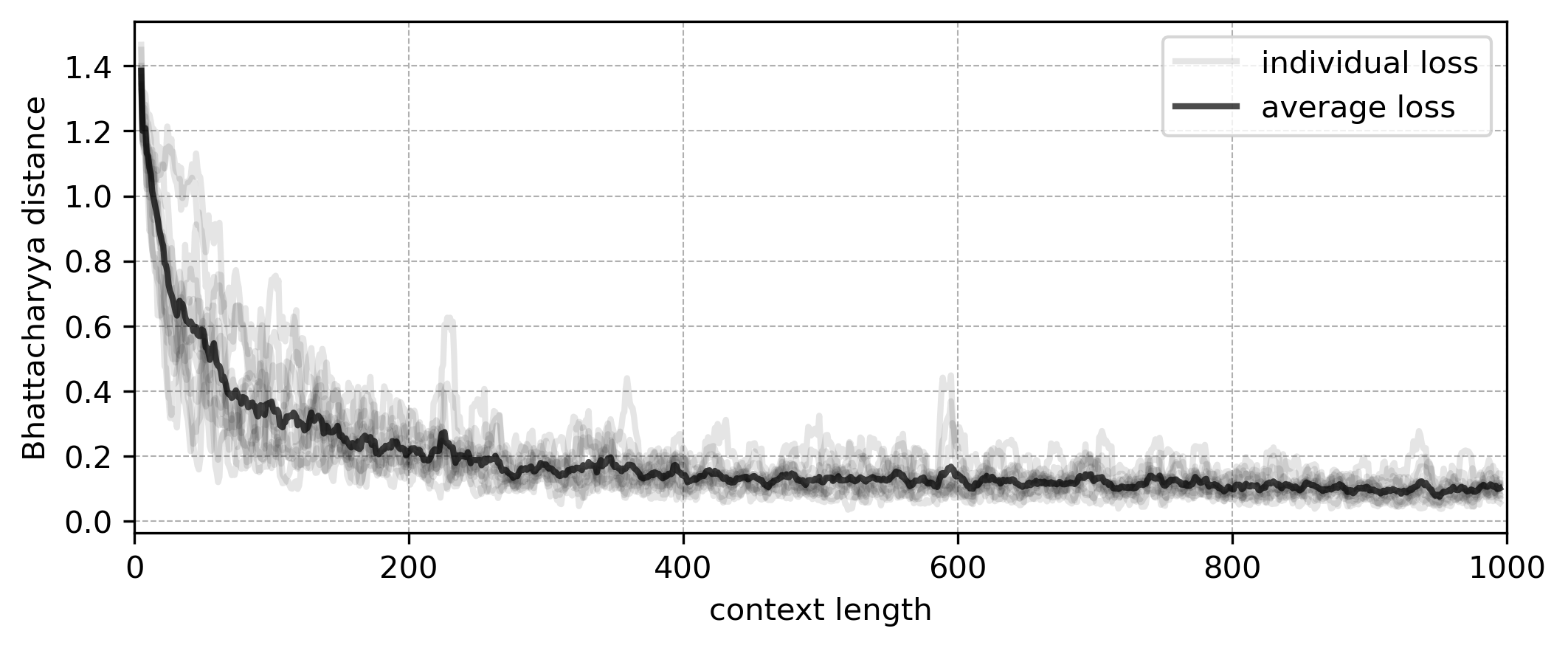}}
    \caption{Stochastic logistic map in-context loss curves.}
    \label{fig:noisy logistic ICL}
  \end{center}
  \vspace{-2em}
  \begin{center}    \centerline{\includegraphics[width=\columnwidth]{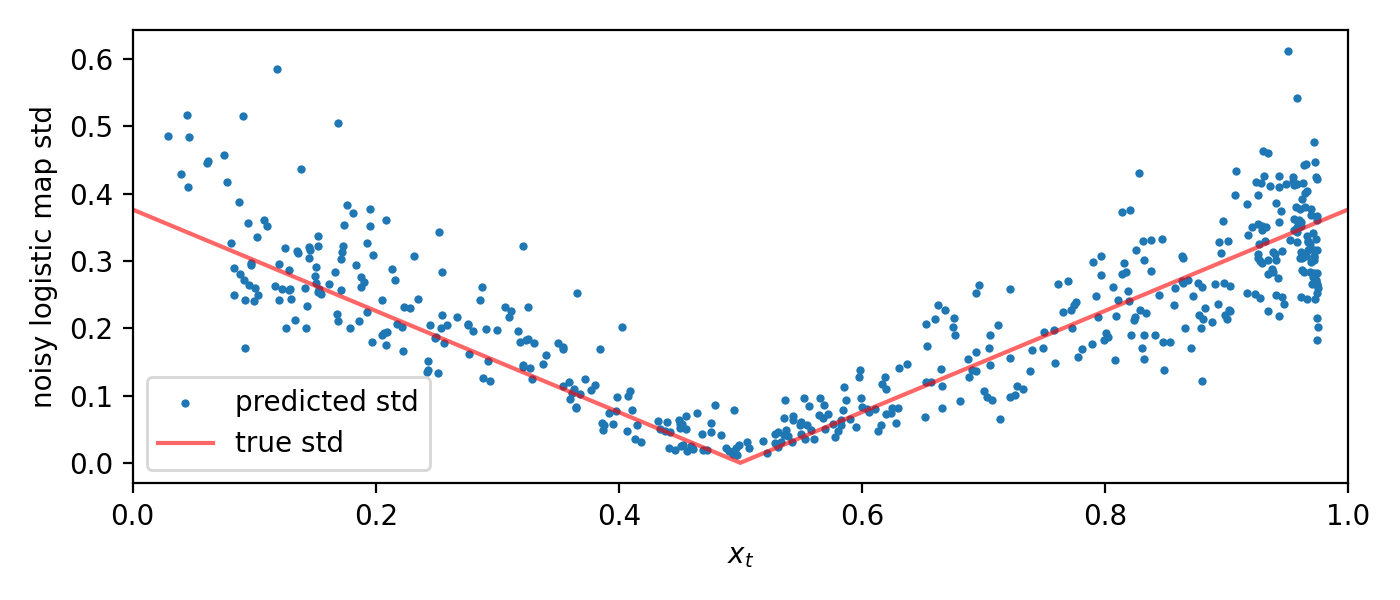}}
    \vspace{-1em}
    \caption{Noisy logistic map standard deviation as a function of the state value $x_t$, learned by the LLM, along with the ground truth.}
    \label{fig:logistic sigma}
  \end{center}
  \vspace{-1.5em}
\end{figure}

We again observe a power-law-like decay of the in-context loss function with respect to the length of the observed time series in \cref{fig:noisy logistic ICL}.
We note that in order to achieve low in-context loss, the LLM must learn to predict not only the mean, but also the variance of next-state distributions. 
This is shown in \cref{fig:logistic sigma} and discussed further in \cref{app:temperature}.

\section{Discussion and conclusion}
\label{sec_conclusions}

\paragraph{Main observations.}
We showed that, with sufficient context, LLMs can accurately recover the probablistic transition rules underlying deterministic, chaotic, and stochastic time series.

\paragraph{In-context neural scaling law.}
Neural scaling laws \cite{kaplan2020scaling} are power laws that characterize how the loss of trained neural networks vary with respect to parameters of the model, such as model size, dataset size, and computational resources.
To the best of our knowledge, neural scaling laws have so far only been observed in the training procedure, which updates the weights of neural networks using an explicit algorithm, such as stochastic gradient descent and Adam \cite{kingma2017adam}.
The loss curves observed in the different numerical experiments (see \cref{fig:ICL,fig:ICL power law}) reveal an additional \emph{in-context} scaling law for LLMs' zero-shot learning of dynamical systems. Further analysis of these scaling laws are presented in \cref{neural_scaling}.

\begin{figure*}[t]
  \centering
  \includegraphics[width=0.49\textwidth]{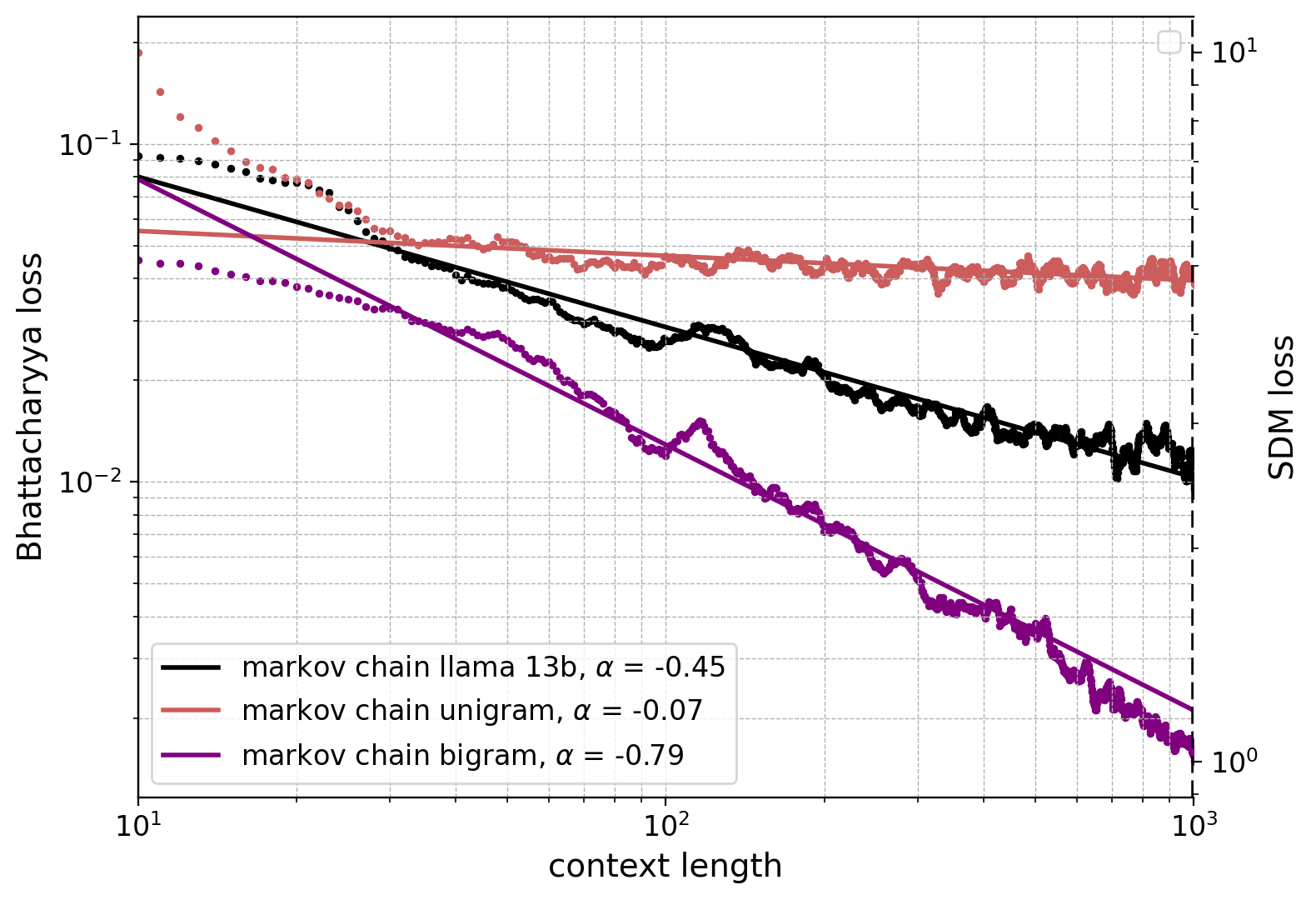}
  \includegraphics[width=0.49\textwidth]{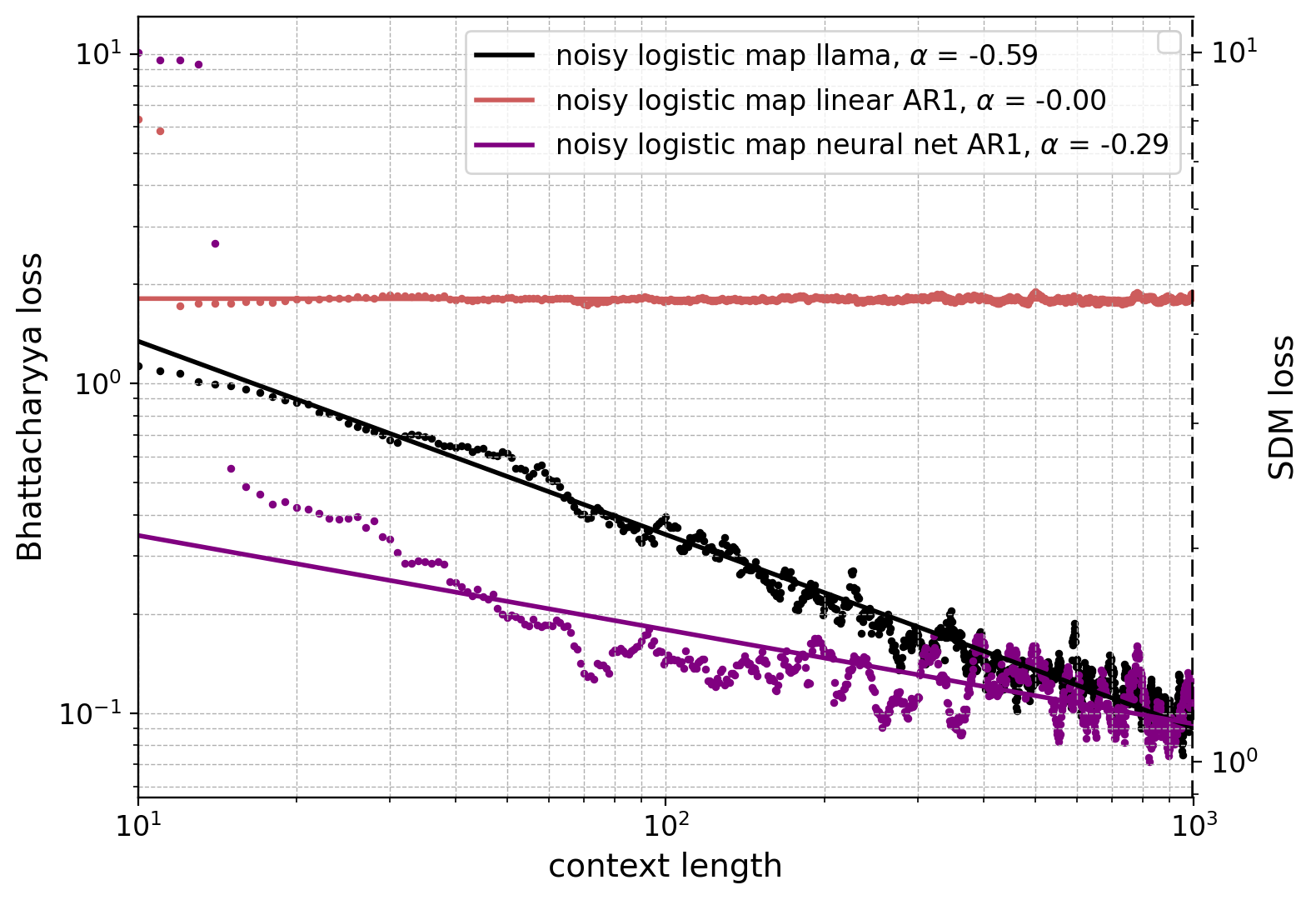}
  \caption{Loss curves of LLM in-context learning against baseline models: unigram and bi-gram models for discrete
  Markov chains, and linear and non-linear autoregressive models with 1-step memory (AR1) for noisy logistic maps.
  The coefficient $\alpha$ denotes the fitted scaling exponent.}
  \label{baseline_plots_sneak_peek}
\end{figure*}

\paragraph{Mechanistic interpretations of in-context learning.}
Characterizing the in-context learning algorithms  implemented by an LLM is an open question of broad interest ~\cite{shen2024revisiting,olsson2022incontextlearninginductionheads}.
Recent years have seen a new paradigm that explains certain aspects of in-context learning as transformers
``hosting" classical algorithms such as gradient descent~\cite{vonoswald2023transformers}.
Our study contributes to this field by presenting novel phenomenological evidence: we demonstrate that pretrained LLMs can in-context learn dynamical systems, 
exhibiting loss curves similar to those of established learning algorithms, such as the bi-gram Markov model and neural network-based autoregressive (AR1) models trained using gradient descent (Figure \ref{baseline_plots_sneak_peek}).
We elaborate on the motivation and design of these baseline models in Appendix \ref{baseline_models}.


\paragraph{Future directions.}
A compelling future direction is to develop a mechanistic theory that explains how pretrained LLMs in-context learn and predict stochastic dynamical time series.
Further investigations could focus on:
\begin{itemize}
    \item Investigating how LLMs encode spatio-temporal features in stochastic dynamical time series \citep{gurnee2023language},
     and how they represent the learned transition rules.
    
    \item Examining how LLMs perform subtasks necessary for learning stochastic dynamical systems, 
    such as collecting Markovian transition statistics from discrete data \citep{akyurek2024incontextlanguagelearningarchitectures} 
    and performing density estimation.

    \item Assessing to what extent training on natural language enhances a transformer's mathematical ability  \citep{zhang2023unveilingtransformerslegosynthetic} to in-context learn dynamical systems,
    and vice-versa.
\end{itemize}


\section*{Limitations}

\paragraph{Data Leakage.}
Although the accurate predictions made by LLMs of the next time step value are highly unlikely to be due to memorization, it is essential to address this possibility thoroughly. Given the vast number of potential sequences even with a thousand numerical values encoded to three digits (resulting in $10^{3000}$ instance), this far exceeds the approximate $10^{12}$ tokens in the training corpus~\cite{touvron2023llama}. Future work should ensure rigorous testing to further rule out data leakage.

\paragraph{Model selection.}
At the time of this manuscript's preparation, newer models, such as LLaMA-3, have been released. Due to constraints in time and computational resources, this work does not evaluate these latest models. Future research should include these and other emerging models to provide a more comprehensive understanding of the phenomena observed.


\section*{Acknowledgements}
This work was supported by the SciAI Center, and funded by the Office of Naval Research (ONR), under Grant Numbers N00014-23-1-2729 and N00014-23-1-2716.

\clearpage
\bibliography{custom}

\newpage
\appendix
\onecolumn
\section{Appendix} \label{Appendix}

\subsection{Loss Functions} \label{loss_functions}

Once the learned transition rules, $\tilde{P}(X_{t+1}|X_t)$, have been extracted, we quantify the deviation from the ground truth $P(X_{t+1}|X_t)$. Depending on the nature of the system, one of the following two loss functions may be more appropriate (see \cref{sec:experiments}).

\paragraph{Bhattacharyya distance.}
For stochastic time series, we use the Bhattacharyya distance to characterize the distance between learned and ground truth transition functions. The Bhattacharyya distance between $P$ and $\tilde{P}$, on a domain $\mathcal{X}$ is defined as~\cite{bhattacharyya1943measure,bhattacharyya1946measure,kailath1967divergence}:
\begin{equation} \label{eq:continuous BT}
  \mathrm{D_B}(P,\tilde P) = -\ln \left(\int_{\mathcal{X}} \sqrt{p(x) \tilde p(x)} \textup{d} x \right),
\end{equation}
and has been widely employed by feature selection and signal extraction methods~\cite{choi2003feature,kailath1967divergence}. Since $\tilde{P}(X_{t+1}|X_t)$ takes the form of a hierarchical PDF, one may approximate the integral in \cref{eq:continuous BT} via a discrete quadrature rule as
\begin{equation} \label{eq:discrete BT}
  \mathrm{D_B}(P,\tilde P) = -\ln \left(\sum_x \sqrt{p(x) \tilde p(x)} \Delta x \right),
\end{equation}
where $\Delta x$ denotes the length of the sub-interval containing $x$ in the partition of $\mathcal{X}$.

\paragraph{Squared deviations from the mean.} For deterministic systems, the true transition functions become delta-functions. As a result, the discretized Bhattacharyya distance from \cref{eq:discrete BT} reduces to (see \cref{eq_DB_NLL} in \cref{app:loss_functions})
\[\mathrm{D_B}(\delta(x-x_\mathrm{true}),\tilde P)=-\frac{1}{2}\ln(\tilde{p}(x_\mathrm{true})) + C,\]
which is proportional to the negative log-likelihood (NLL) assigned to the true data by the LLM, plus a constant $C$\footnote{This constant is determined by the base B of the system, and the number of digits n as $C=-\ln \Delta x = n \log B$.}. NLL references only the finest bins in the hierarchical PDF and is thus unstable as an in-context loss.
As an alternative, we use the squared deviations from the mean (SDM)~\cite{kobayashi2000comparing} as the in-context loss for deterministic systems:
\[
  \mathrm{SDM}(x_{\mathrm{true}}, \tilde{P})
  = \left(x_{\mathrm{true}} - \sum_{x \in \mathcal{X}} \tilde{p}(x) x \Delta x\right)^2,
\]
where the mean $\mu_{\tilde{P}}=\sum_x \tilde{p}(x) x \Delta x$ is extracted from the hierarchical PDF. Note that unlike the Bhattacharyya distance, which references the model prediction $\tilde{p}$ only at $x_\mathrm{true}$, the SDM takes into account the entire support $x \in \mathcal{X}$. Our numerical experiments suggest that SDM is more stable and better captures the in-context learning dynamics of deterministic systems (see \cref{additional_experiments_deterministic}).

\subsection{Additional loss functions} \label{app:loss_functions}
\paragraph{KL-divergence.}
The KL-divergence between two PDFs, $P$ and $\tilde{P}$, is defined as
\begin{equation} \label{eq_KL}
  D_{KL}(P,\tilde{P}) = \sum_{x\in\mathcal{X}}P(x) \log \left(\frac{P(x)}{\tilde{P}(x)}\right).
\end{equation}
Although commonly used as the training loss for a variety of machine learning systems, this loss function may suffer from numerical instabilities as the learned transition function $\tilde{P}$ are often close to zero, as shown in \cref{fig:logistic example,fig:BM example}, where the probability density is concentrated in small regions of the support.

\paragraph{Discretized Bhattacharyya distance for deterministic systems.}
For deterministic systems, the ground truth transition function is a delta function. Therefore, the Bhattacharyya distance between it and the Hierarchy-PDF prediction only references the finest bin associated with the true value $x_{\mathrm{true}}$.
\begin{equation} \label{eq_DB_NLL}
  \begin{aligned}
    D_B(\delta(x-x_\mathrm{true}),\tilde P)
     & = -\ln \left(\sum_x \sqrt{\delta (x-x_\mathrm{true}) \tilde p(x)} \Delta x \right)
    = -\frac{1}{2}\ln(\tilde{p}(x_\mathrm{true}))-\ln \Delta x                            \\
     & = -\frac{1}{2}\ln(\tilde{p}(x_\mathrm{true})) + \text{constant}.
  \end{aligned}
\end{equation}
As a result, the Bhattacharyya distance is reduced to an affine-transformed negative log-likelihood assigned to data by the LLM. Such local sensitivity on $\tilde{p}(x_\mathrm{true})$ explains the wild fluctuations seen in the Bhattacharyya loss in \cref{additional_experiments_deterministic}.

\begin{figure}[htbp]
  \begin{center}
    \centerline{\includegraphics[width=0.8\columnwidth]{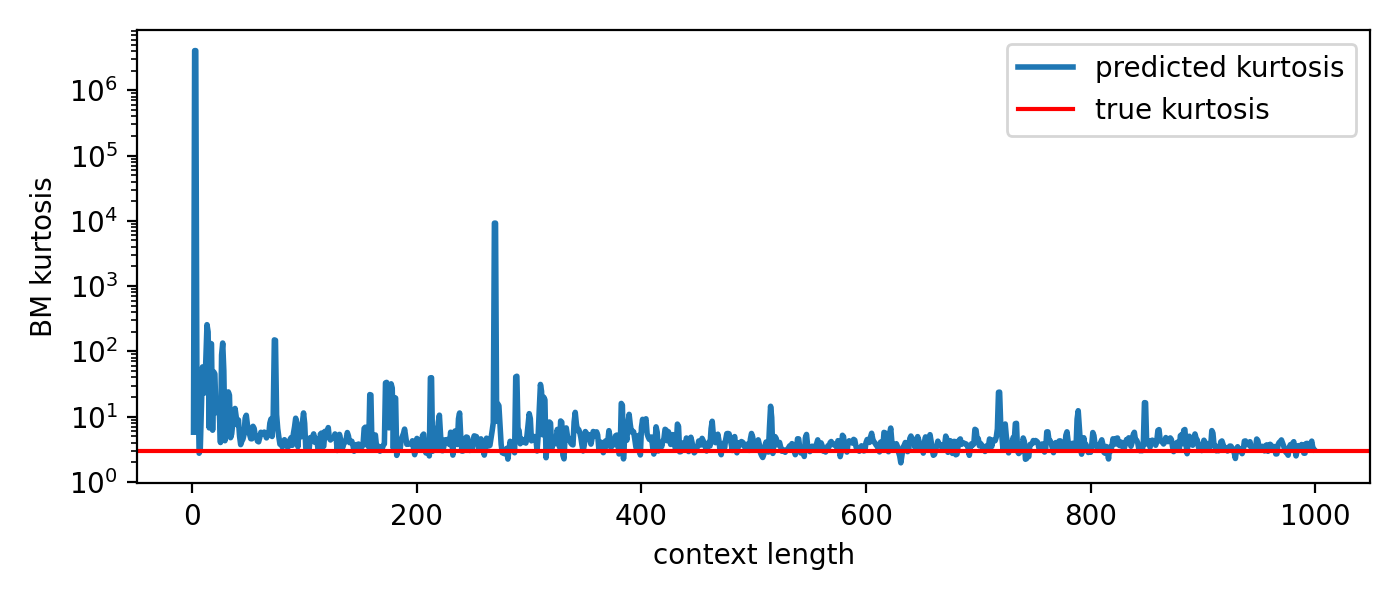}}
    \vspace{-1em}
    \caption{Kurtosis of Brownian motion transition rules with respect to the input length. Blue: kurtosis
      of LLM predicted PDF. Red: ground truth kurtosis, which is $3$ for all Gaussian distributions.}
    \label{fig: BM kurtosis}
  \end{center}
  \vspace{-2em}
\end{figure}

\paragraph{Higher moments and kurtosis.}
While the Bhattacharyya distance and SDM measure the agreement between the extracted transition rules $\tilde{P}$ and the ground truth distribution $P$, they do not explicitly characterize the type of the distribution (e.g., Gaussian or uniform). We employ the kurtosis as an additional measure to assess whether the LLM recovers the correct shape of $P$. The kurtosis of a distribution $P$ is defined as~\cite{joanes1998comparing}
\begin{equation}
  \text{Kurt}(P) = \frac{\mathbb{E}_{x\sim P} [(x-\mu_P)^4]}{\mathbb{E}_{x\sim P} [(x-\mu_P)^2]^2} = \frac{\mu_4}{(\sigma^2)^2},
\end{equation}
where $\sigma^2$ and $\mu_4$ are the second and fourth central moments, which can be approximated using a hierarchical PDF as
\begin{equation} \label{eq:PDF moments}
  \sigma^2(P) = \sum_x p(x)(x-\mu_p)^2 \Delta x,\quad
  \mu_4(P)    = \sum_x p(x)(x-\mu_p)^4 \Delta x.
\end{equation}
The kurtosis is equal to $3$ for a Gaussian distribution and $9/5$ for bounded uniform distributions. \cref{fig: BM kurtosis} shows the kurtosis of Brownian motion transition rules learned by LLM, which converges to $3$ as the context length increases.

\subsection{Additional Experiments: stochastic time series} \label{additional_experiments_stochastic}

\subsubsection{Brownian motion}

Brownian motion is an example of a continuous-time stochastic process
\cite{einstein1905molekularkinetischen}, and is described by a stochastic differential equation (SDE):
\begin{equation} \label{eq_brownian}
  dX_t = \mu dt + \sigma dW_t,
\end{equation}
where $X_t$ represents the state of the system at time $t$, $\mu$ is the drift coefficient, $\sigma$ is the volatility coefficient, and $dW_t$ is the increments of a Wiener process~\cite{revuz2013continuous}, modeling the randomness of motion.

\begin{figure}[htbp]
  \begin{center}
    \centerline{\includegraphics[width=0.7\textwidth]{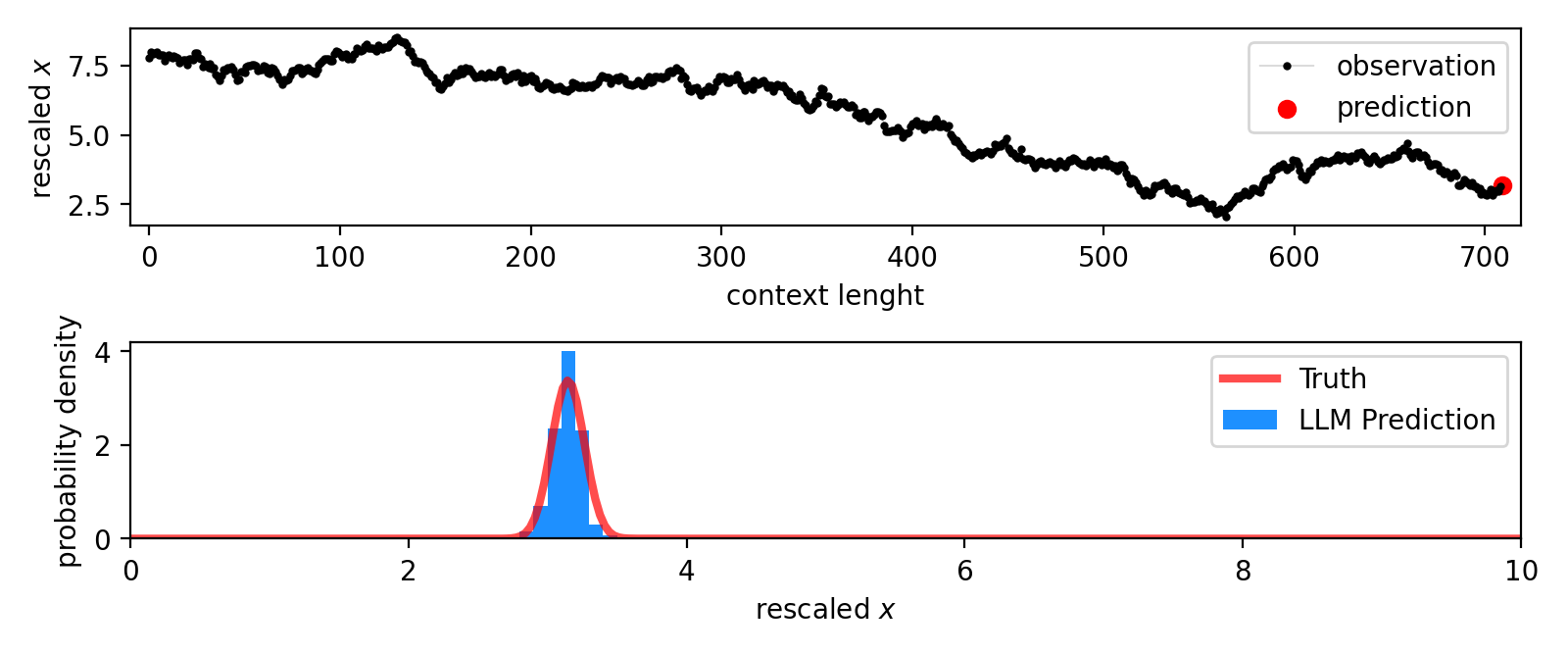}}
    \vspace{-1em}
    \caption{Next state prediction of Brownian motion.
      Top: Input stochastic time series shown in black, and the state to be predicted is highlighted in red. Bottom: The LLM's prediction, along with the ground truth distribution.}
    \label{fig:BM example}
  \end{center}
  \vspace{-2em}
\end{figure}

\begin{figure}[htbp]
  \begin{center}
    \centerline{\includegraphics[width=0.7\columnwidth]{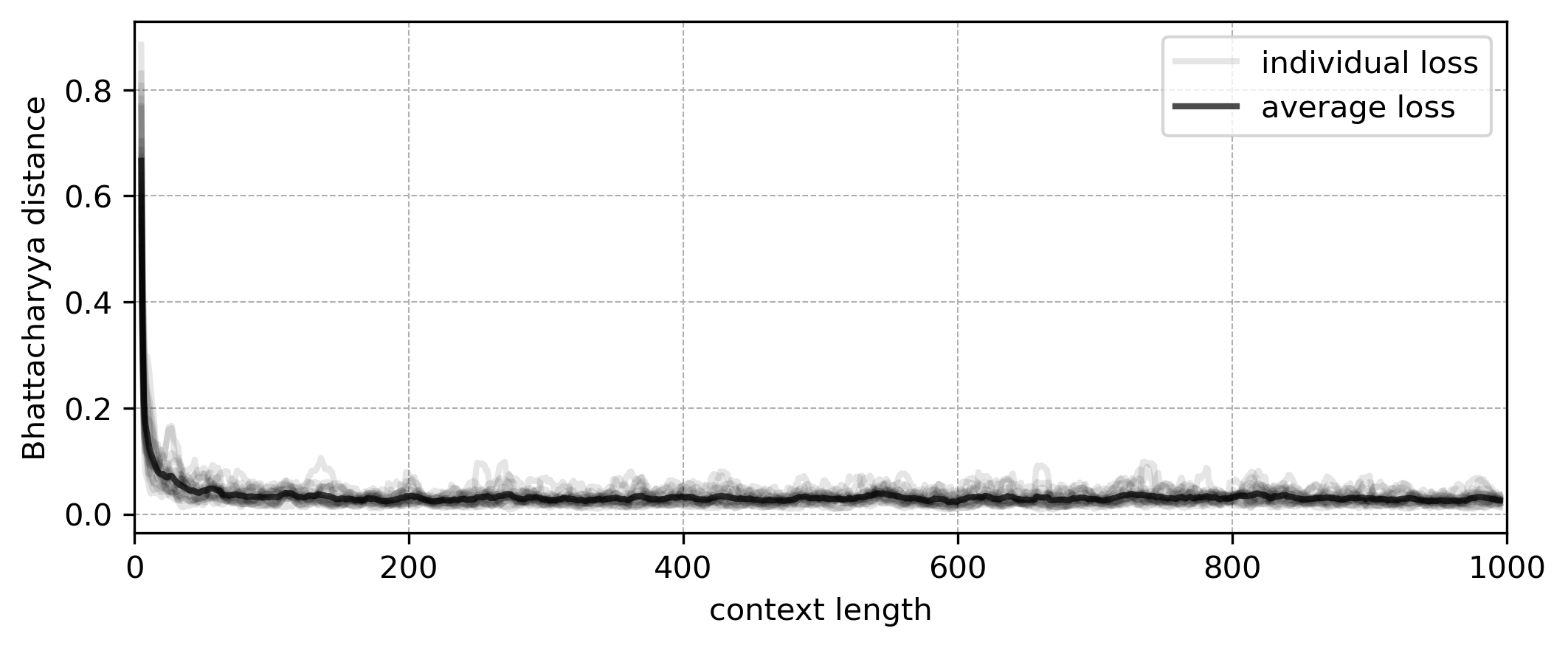}}
    \vspace{-1em}
    \caption{Bhattacharyya distance between the LLM predicted PDF and the ground truth transition function of Brownian motion with respect to the input length.}
    \label{fig:BM ICL}
  \end{center}
  \vspace{-2.5em}
\end{figure}

To simulate trajectories of Brownian motion, we use the Euler--Maruyama method~\cite{platen1999introduction}, which discretizes \cref{eq_brownian} as $X_{t+\Delta t} = X_t + \mu \Delta t + \sigma \sqrt{\Delta t} Z$, where $\Delta t$ is the time resolution, and $Z\sim \mathcal{N}(0,1)$ is a random variable that follows a standard Gaussian distribution. The Euler--Maruyama method may also be written as a conditional distribution:
\[
  X_{t+\Delta t}|\{X_t=x_t\} \sim \mathcal{N}(x_t+\mu\Delta t, \sigma^2 \Delta t),
\]
which is the ground truth transition function visualized in \cref{fig:BM example}. Indeed, the ground truth next state is described as a Gaussian distribution, and we observe in \cref{fig:BM example} that the LLM prediction agrees well with the true, underlying distribution. Additionally, as shown in \cref{fig:BM example}, the LLM displays the correct Gaussian shape for the PDF, converging to a measured kurtosis of $3$ (see \cref{app:loss_functions}). We then simulate ten different trajectories using random seeds for $Z$ and report the resulting LLM learning curves in \cref{fig:BM ICL}, measured in the Bhattacharyya distance.

\subsubsection{Geometric Brownian motion}

\begin{figure}[ht]
  \begin{center}    \centerline{\includegraphics[width=0.7\textwidth]{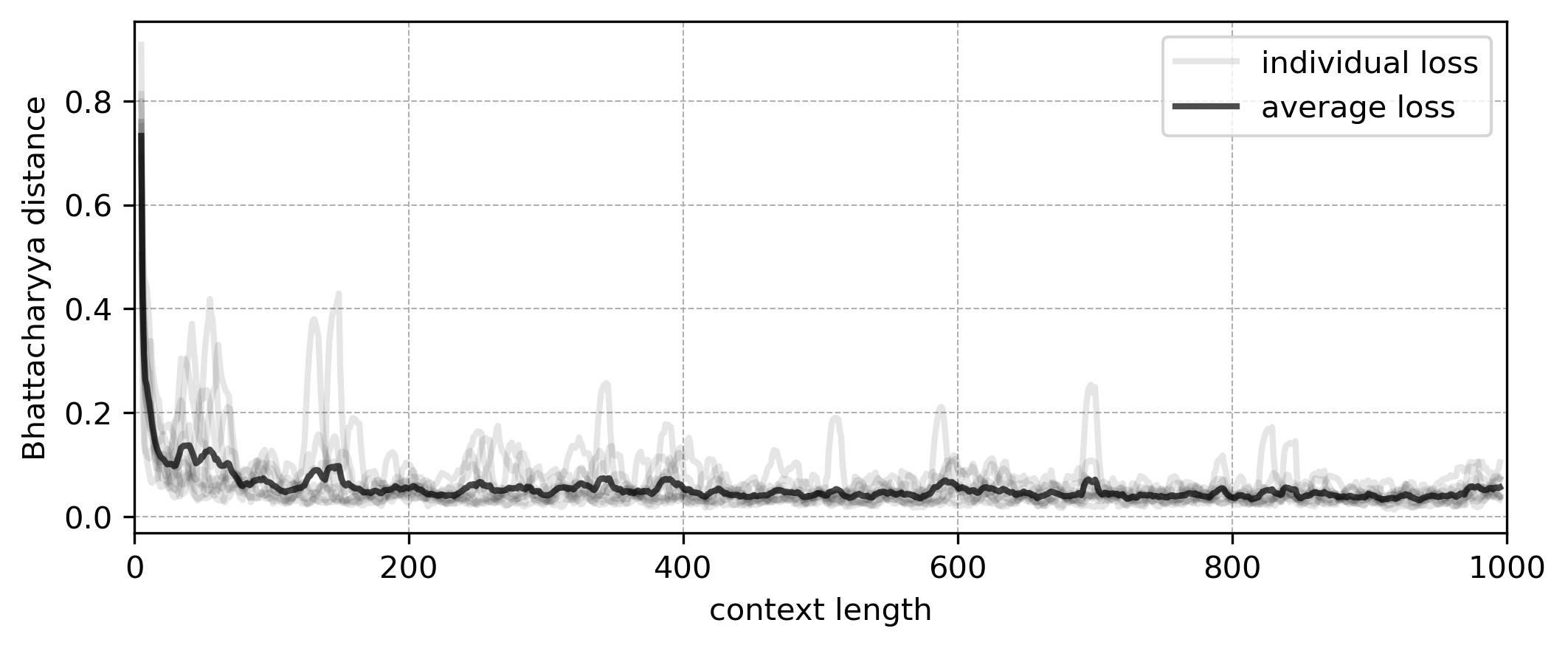}}
    \vspace{-1em}
    \caption{Geometric Brownian motion in-context loss curve.}
    \label{fig:GBM ICL}
  \end{center}
  \vspace{-2em}
\end{figure}

Geometric Brownian motion (GBM)~\cite{oksendal2013stochastic} is a stochastic process that is commonly used in mathematical finance to model the trajectories of stock prices and other financial assets~\cite{hull2003options}. A GBM is governed by the following SDE:
\begin{equation} \label{GBM_SDE}
  dX_t = \mu X_t dt + \sigma X_t dW_t,
\end{equation}
where $X_t$ models the price of an asset at time $t$, and the fluctuation term $\sigma X_t dW_t$ is proportional to the current asset price $X_t$. The Euler--Maruyama discretization of the GBM reads $X_{t+\Delta t} = X_t + \mu X_t \Delta t + \sigma X_t \sqrt{\Delta t} Z$, and leads to the ground truth transition function:
\begin{equation} \label{eq:GBM transition function}
  X_{t+\Delta t}|\{X_t=x_t\} \sim \mathcal{N}(x_t+\mu x_t \Delta t, (\sigma x_t)^2 \Delta t).
\end{equation}
We simulate ten different GBM trajectories using random seeds and report the corresponding learning curves in \cref{fig:GBM ICL}.

\begin{figure}[ht]
  \begin{center}    \centerline{\includegraphics[width=0.7\textwidth]{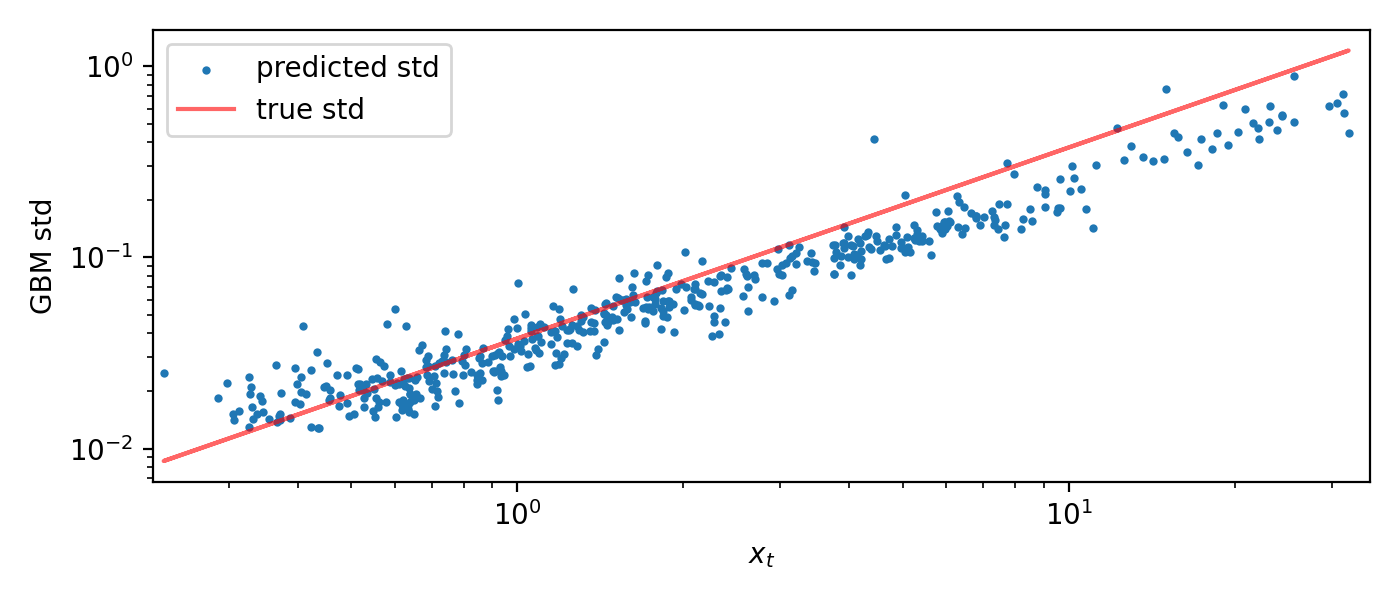}}
    \vspace{-1em}
    \caption{Evolution of the geometric Brownian motion standard deviation with respect to the state value $x_t$ (see \cref{eq:GBM transition function}), along with the predicted standard deviation extracted from the LLM at each time step.}
    \label{fig:GBM sigma}
  \end{center}
  \vspace{-2em}
\end{figure}

We perform an additional numerical test to verify that the LLM is learning the correct relationship between the variance of the GBM and the state value $X_t$ (see \cref{eq:GBM transition function}). To investigate this, we display in \cref{fig:GBM sigma} the expected standard deviation along with the learned one, extracted from the Hierarchy-PDF using \cref{eq:PDF moments}, across all predicted states. We find that the LLM respects the ground truth standard deviation of the GBM, as prescribed by the underlying transition function.

\subsubsection{Markov chains with LLaMA-70b} \label{Markov chains with larger numbers of states}
\begin{figure}[htbp]
  \centering
  \includegraphics[width=0.49\textwidth]{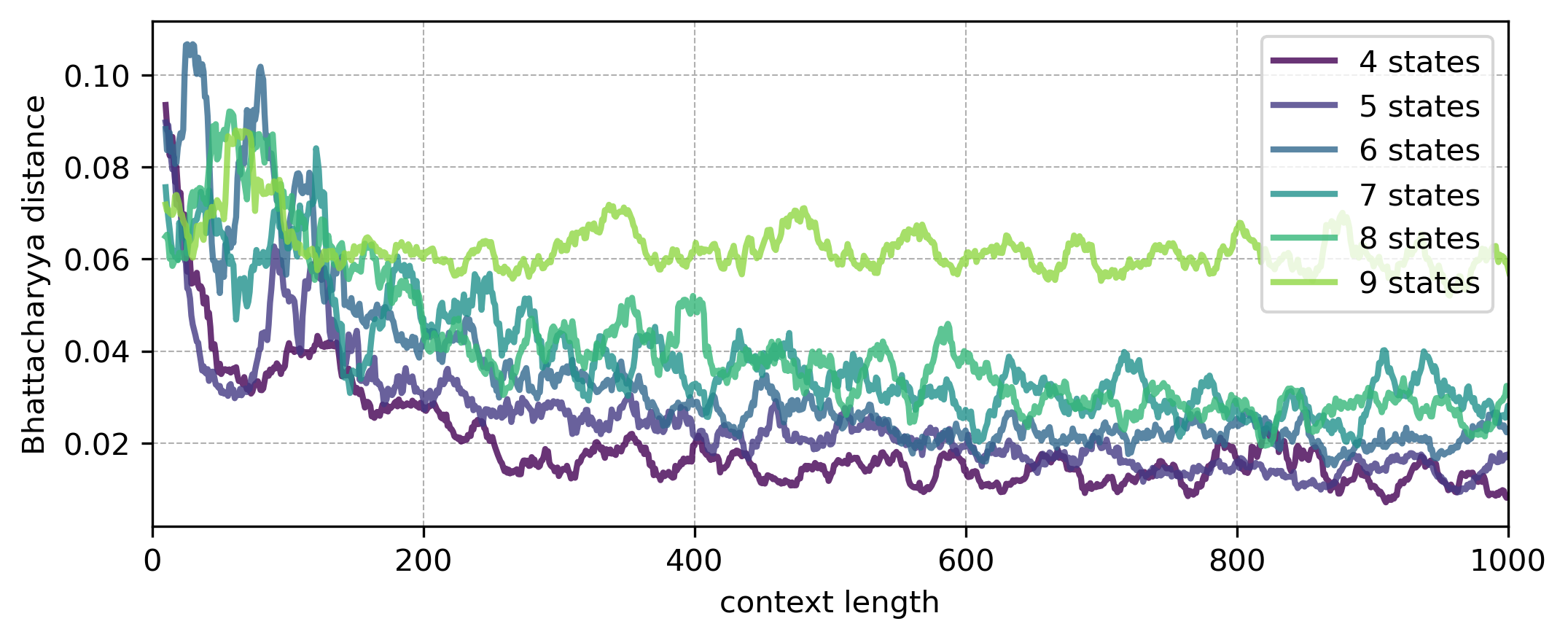}
  \includegraphics[width=0.49\textwidth]{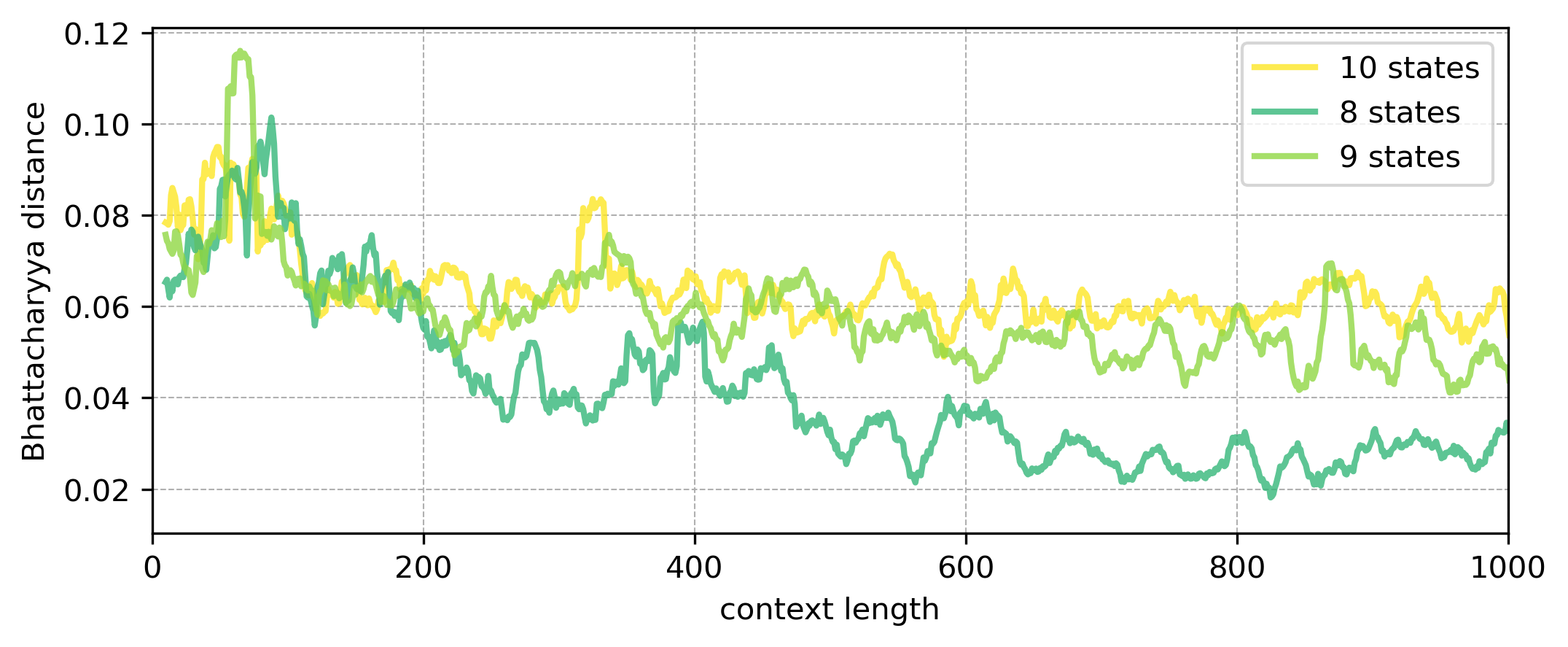}
  \caption{In-context loss curves for LLaMA-13b (left) and LLaMA-70b (right) with respect to the number of states in the transition matrix.}
  \label{fig:markov_chain_13b}
\end{figure}

Our experiments show that LLMs generally achieve lower in-context loss for Markov chains with fewer discrete states $n$, as shown in \cref{fig:markov_chain_13b}. For both LLaMA-13b and LLaMA-70b, the in-context loss curves cease to decrease significantly for numbers of states $n\geq 9$.

\subsection{Additional Experiments: deterministic time series} \label{additional_experiments_deterministic}

\subsubsection{Logistic map}
\label{sec:logistic map}
\begin{figure}[htbp]
  \begin{center}
    \includegraphics[width=0.7\textwidth]{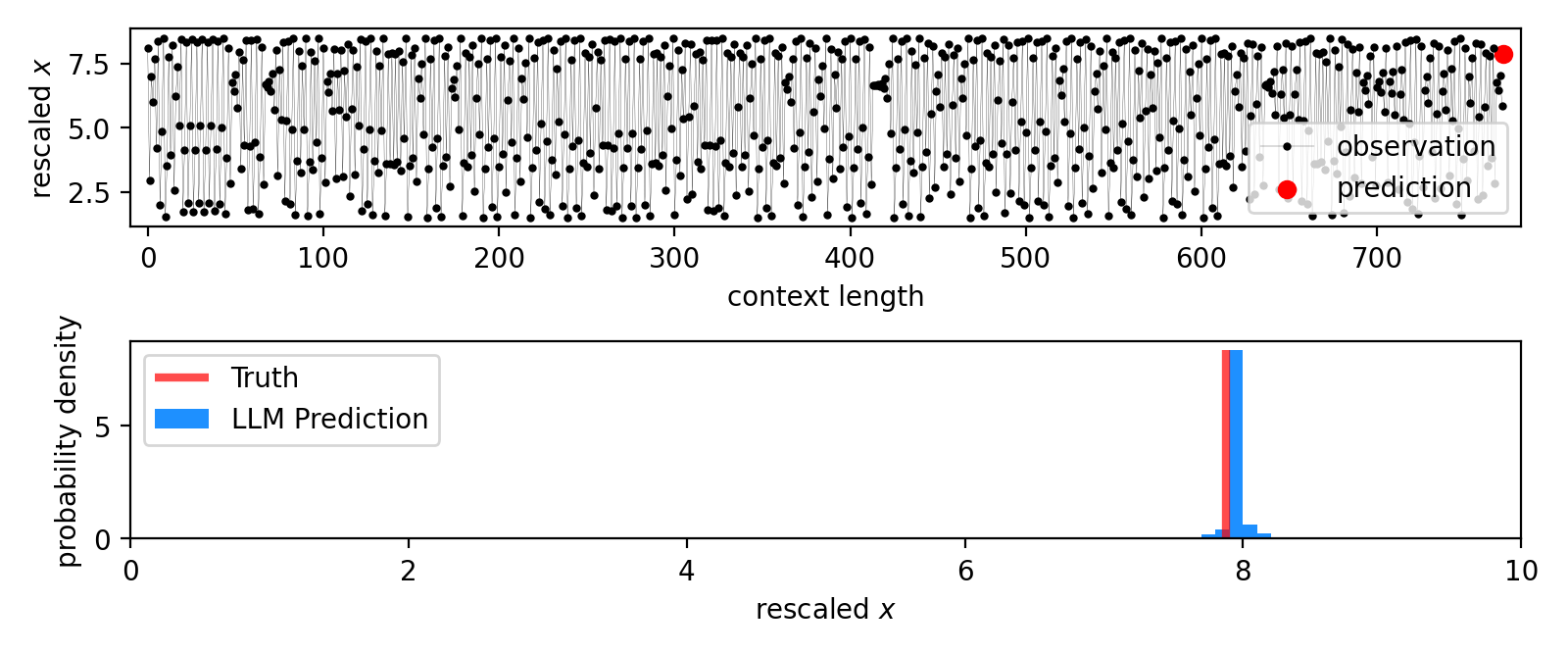}
    \vspace{-1em}
    \caption{Next state prediction of the logistic map. Top: Input chaotic time series shown in black, and the state to be predicted is highlighted in red. Bottom: The LLM's statistical prediction for the last state. The ground truth distribution is delta-distributed, which is shown as a vertical red line.}
    \label{fig:logistic example}
  \end{center}
  \vspace{-1em}
\end{figure}

The logistic map, first proposed as a discrete-time model for population growth, is one of the simplest dynamical systems that manifest chaotic behavior~\cite{strogatz2018nonlinear}. It is governed by the following iterative equation:
\begin{equation} \label{eq: logistic iterative}
  x_{t+1} = f(x_t) = r x_t(1-x_t),\quad x_0 \in (0,1),
\end{equation}
which may also be written using conditional distributions to reflect the deterministic nature of the system as $X_{t+1}|\{X_t=x_t\} \sim \delta_{f(x_t)}$, where $\delta$ denotes the Dirac delta distribution. This conditional distribution is the ground truth transition function displayed in red in \cref{fig:logistic example}. The parameter $r \in [1,4)$ controls the behavior of the system and is set to $r=3.9$. At this value, the dynamics are naturally confined within the interval $(0,1)$, and the system has no stable fixed points. Due to the chaotic nature of the system, two initial nearby trajectories diverge exponentially in time. This property allows us to sample multiple uncorrelated trajectories by using different initial conditions, $x_0$, sampled uniformly in $(0,1)$.

\begin{figure}[ht]
  \begin{center}
    \centerline{\includegraphics[width=0.7\textwidth]{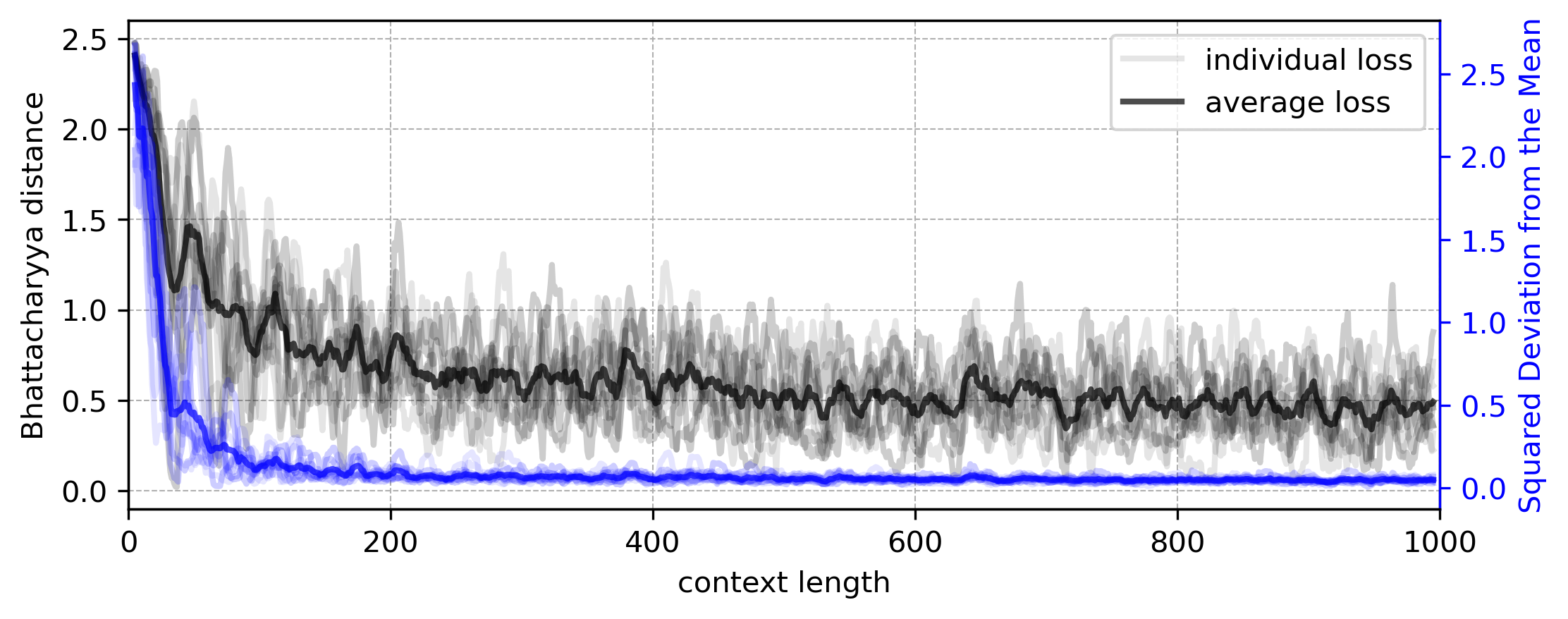}}
    \vspace{-1em}
    \caption{Logistic map in-context loss curves. For deterministic systems, Bhattacharyya loss is subject to large fluctuations while SDM loss is more stable.}
    \label{fig: logistic}
  \end{center}
  \vspace{-2em}
\end{figure}

\cref{fig:logistic example} displays one of the ten tested trajectories and an LLM's prediction of the last state. The PDF of the next state prediction is extracted using the Hierarchy-PDF algorithm described in \cref{sec: extract transition rules}. We find that the LLM prediction is close to the ground truth, except for minor deviations manifested by small, but non-zero, probability densities in neighboring values. While the extracted prediction is only reported for the last time step in the bottom panel of \cref{fig:logistic example}, we also extract the model prediction at every time step for all tested trajectories and report the corresponding Bhattacharyya and SDM losses in \cref{fig: logistic}.

As foreshadowed in \cref{sec:methods}, the Bhattacharyya loss suffers from large fluctuations with deterministic systems such as the logistic map, while the SDM loss better captures the in-context learning dynamics. In particular, the SDM loss decreases rapidly with the number of observed states, without any fine-tuning nor prompt engineering of the LLM. This suggests that the LLM can extract the underlying transition rules of the logistic map from in-context data.

\subsubsection{Lorenz system}

The Lorenz system~\cite{lorenz1963deterministic} is a three-dimensional (3D) dynamical system derived from a simplified model of convection rolls in the atmosphere. It consists of a system of three ordinary differential equations:
\[
  \dot{x}(t) = \sigma(y - x),\quad \dot{y}(t) = x(\rho - z) - y, \quad
  \dot{z}(t) = xy - \beta z,
\]
where $\sigma=10$, $\rho=28$ and $\beta=8/3$ are parameters dictating the chaotic behavior of the system. We compute ten 3D trajectories using a first-order explicit time-stepping scheme. All trajectories share the same initial conditions in $y$ and $z$, and differ only in the $x$-coordinate, which is uniformly sampled in $(0,0.3)$. The chaotic nature of the system guarantees that the sampled trajectories quickly diverge from one another. We prompt the LLM with the $x$-component of the simulated series and extract the next predicting values.

\begin{figure}[htbp]
  \begin{center}
    \centerline{\includegraphics[width=0.7\textwidth]{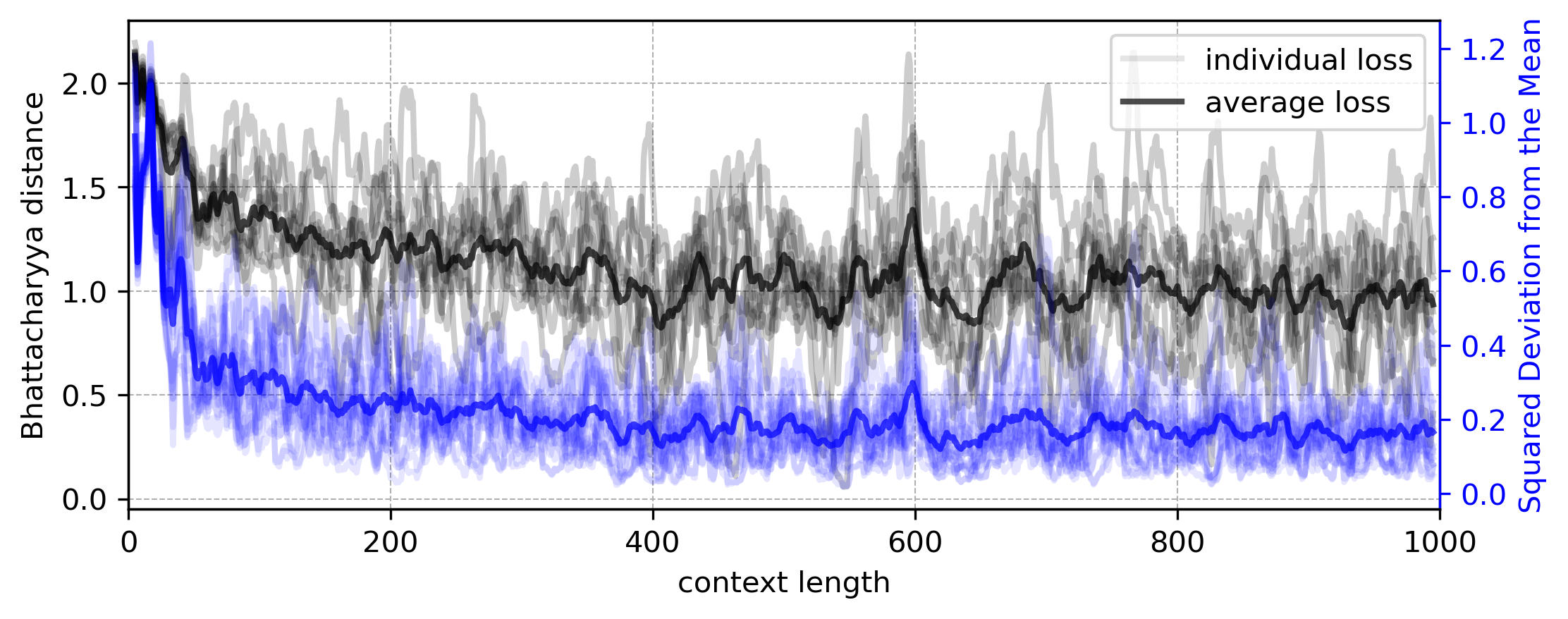}}
    \vspace{-1em}
    \caption{Loss curves for predicting the $x$-component of the Lorenz system with respect to the number of observed time steps.}
    \label{fig:lorenz ICL}
  \end{center}
  \vspace{-2em}
\end{figure}

When the $x$, $y$, and $z$ components are observed, the system is deterministic and Markovian; in the sense that a state vector $\vec{s}_t=(x_t,y_t,z_t)$ at time $t$ fully determines the next state $\vec{s}_{t+1}$. However, if the $x$-component is the only one observed, then the system ceases to be Markovian but remains deterministic if one expands the state vector to include information from earlier states. Hence, Takens' embedding theorem~\cite{takens2006detecting} guarantees that the observation of at most seven states of the series $x_{0:t}$ is sufficient to predict $x_{t+1}$. Finding the optimal number of states to reconstruct the system's trajectory is an area of active research~\cite{strogatz2018nonlinear}. Despite this apparent difficulty, LLaMA-13b can formulate increasingly accurate predictions of the series as it observes more states, as evidenced by the decaying loss curves plotted in \cref{fig:lorenz ICL}.

\subsection{Continuous State Space Visualization} \label{continuous_visualization}
\begin{figure}[htbp]
  \centering
  \begin{minipage}[b]{0.47\textwidth}
    \centering
    \includegraphics[width=\textwidth]{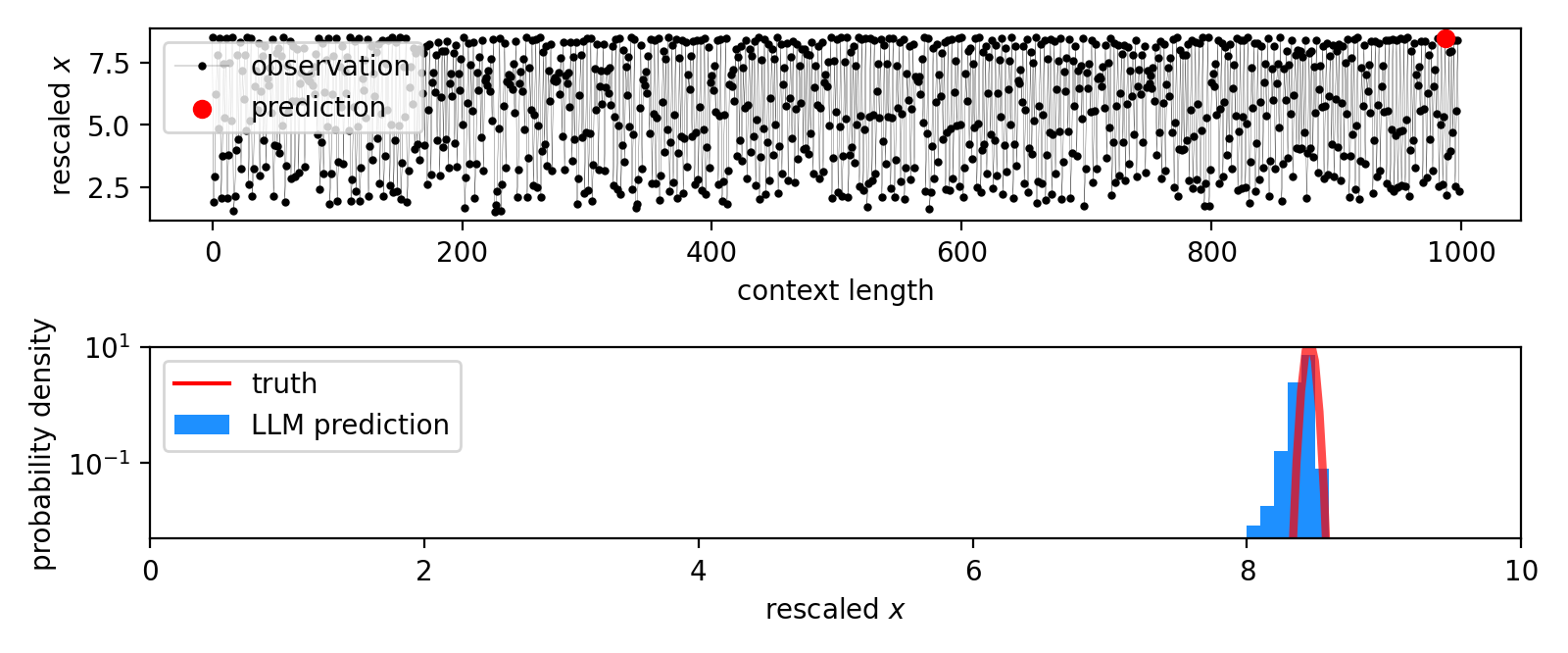}
    \includegraphics[width=\textwidth]{figures/noisy_logistic_example_snapshot_988.png}
    \includegraphics[width=\textwidth]{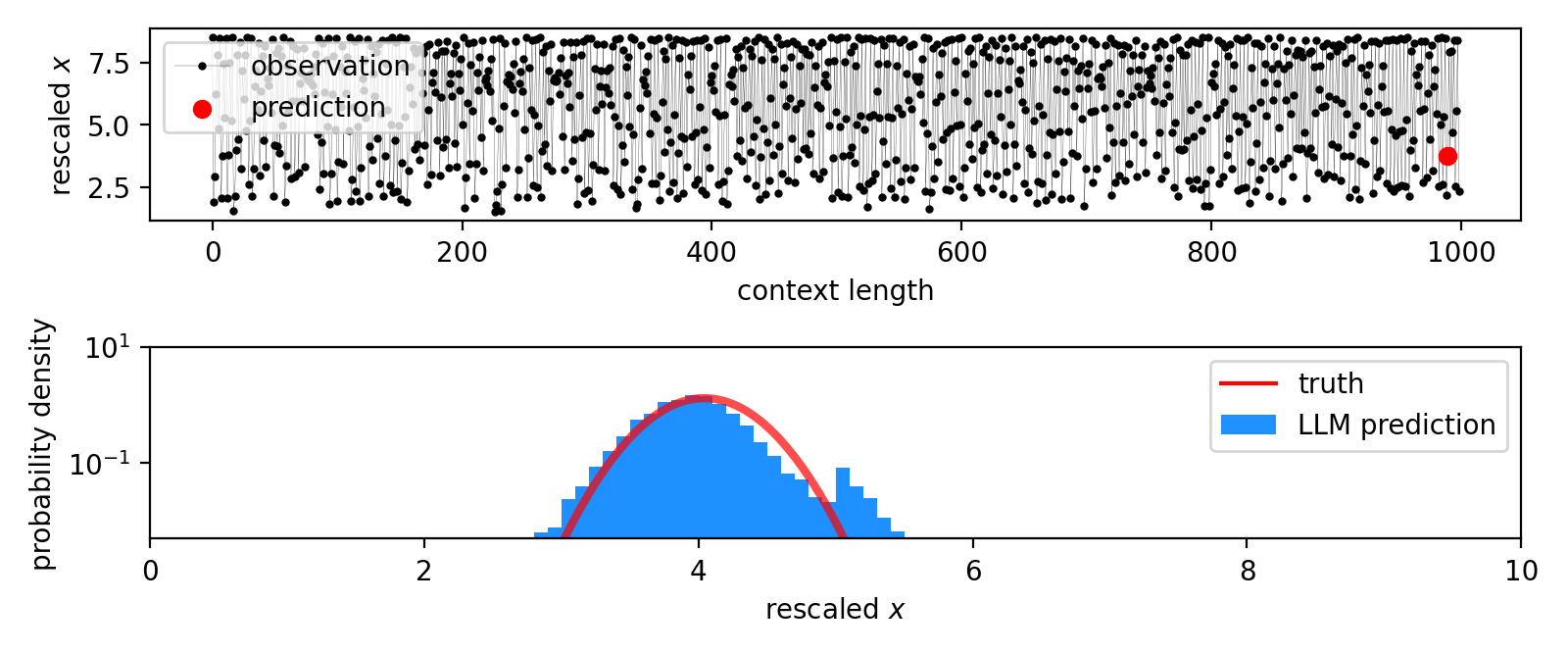}
    \includegraphics[width=\textwidth]{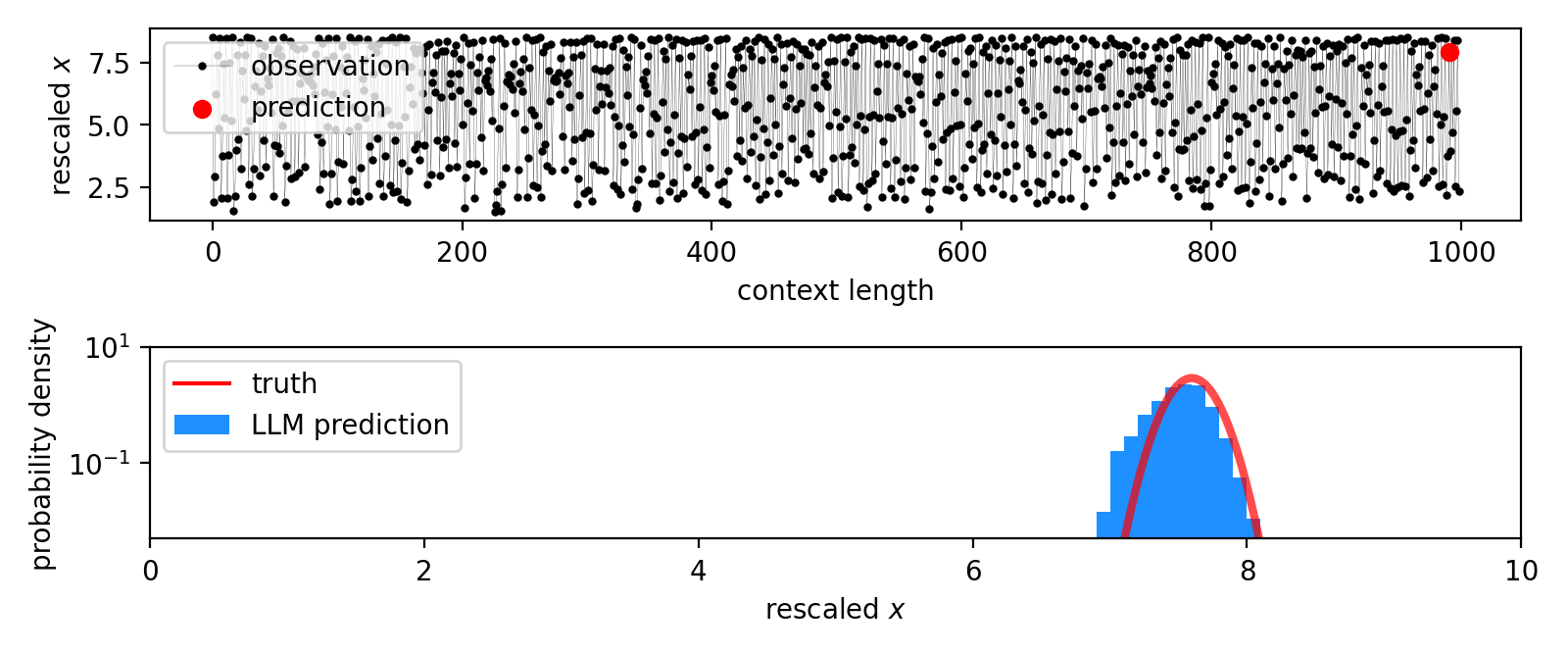}
    \caption{4 consecutive states in a noisy logistic map. Ground truth is shown in red and LLaMA predictions in blue.}
    \label{noisy_logistic_map_snapshots}
  \end{minipage}%
  \hspace{0.04\textwidth} 
  \begin{minipage}[b]{0.47\textwidth}
    \centering
    \includegraphics[width=\textwidth]{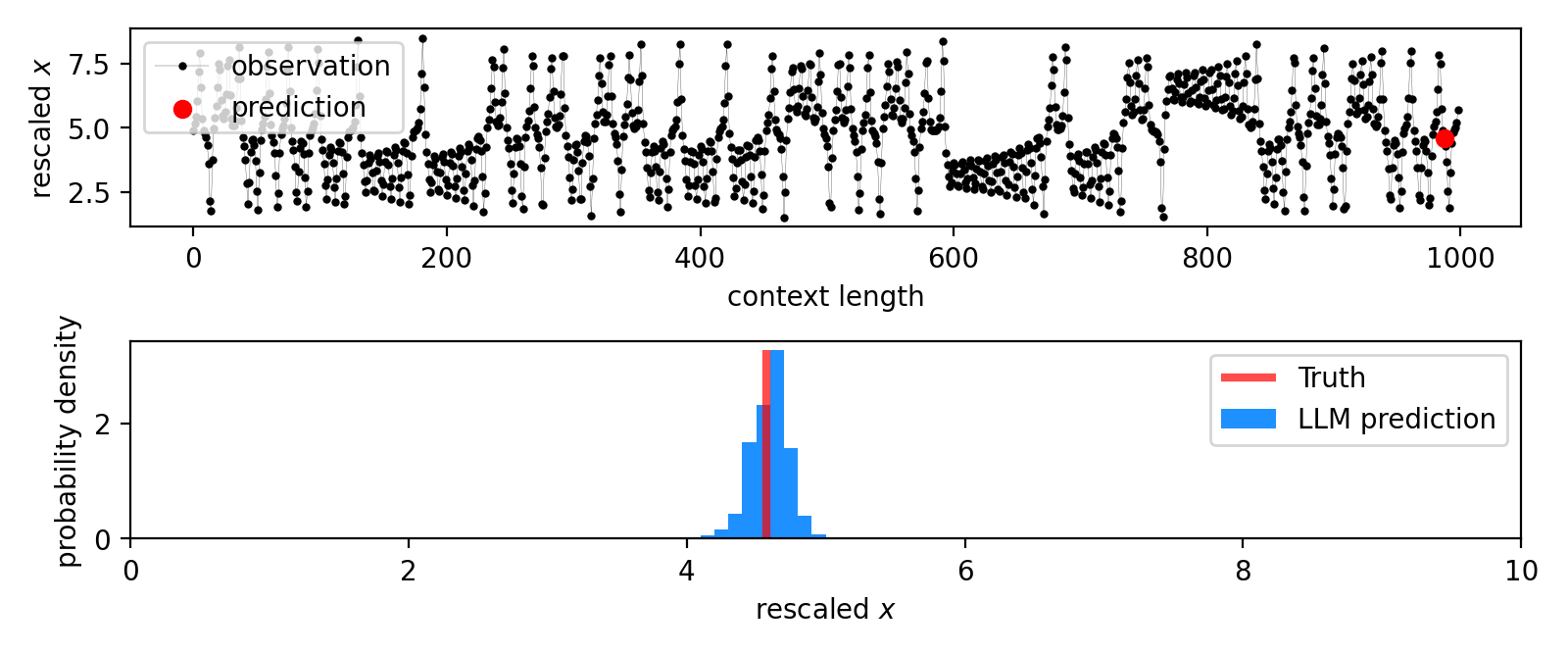}
    \includegraphics[width=\textwidth]{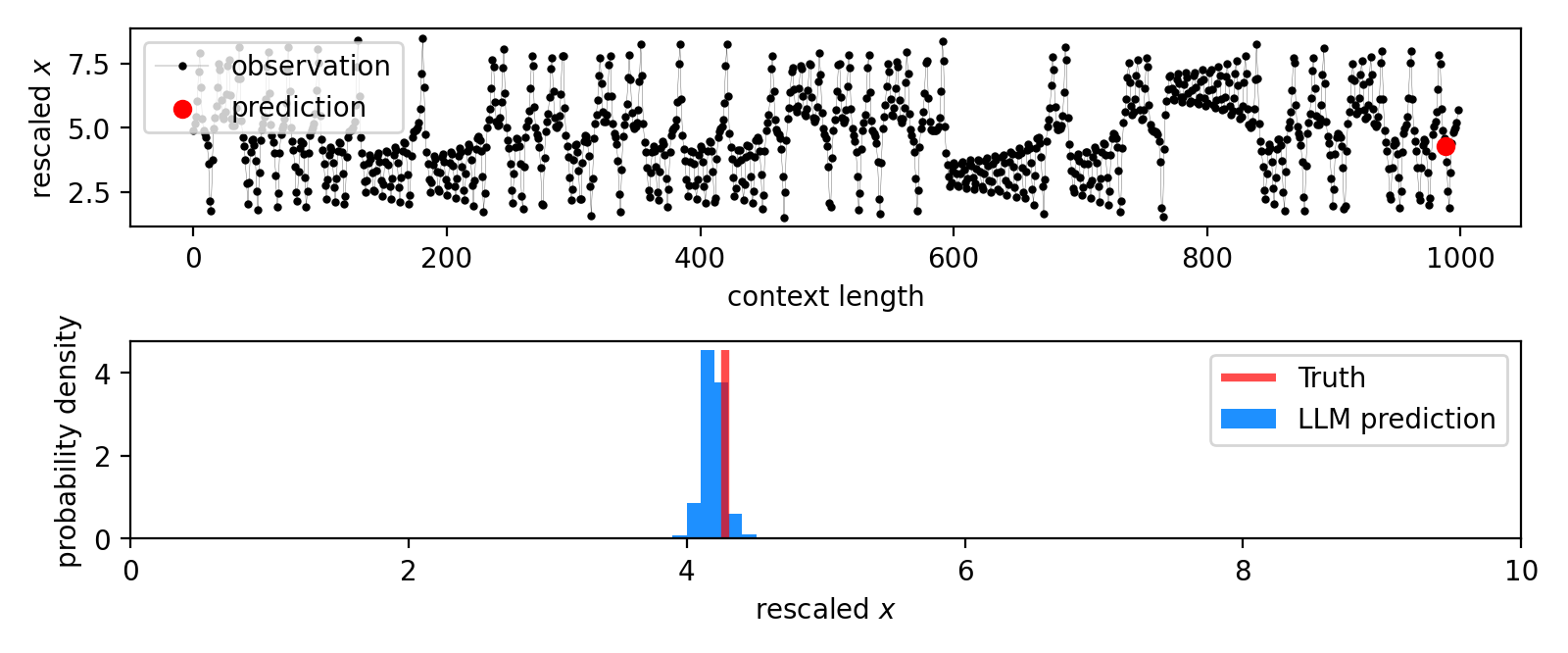}
    \includegraphics[width=\textwidth]{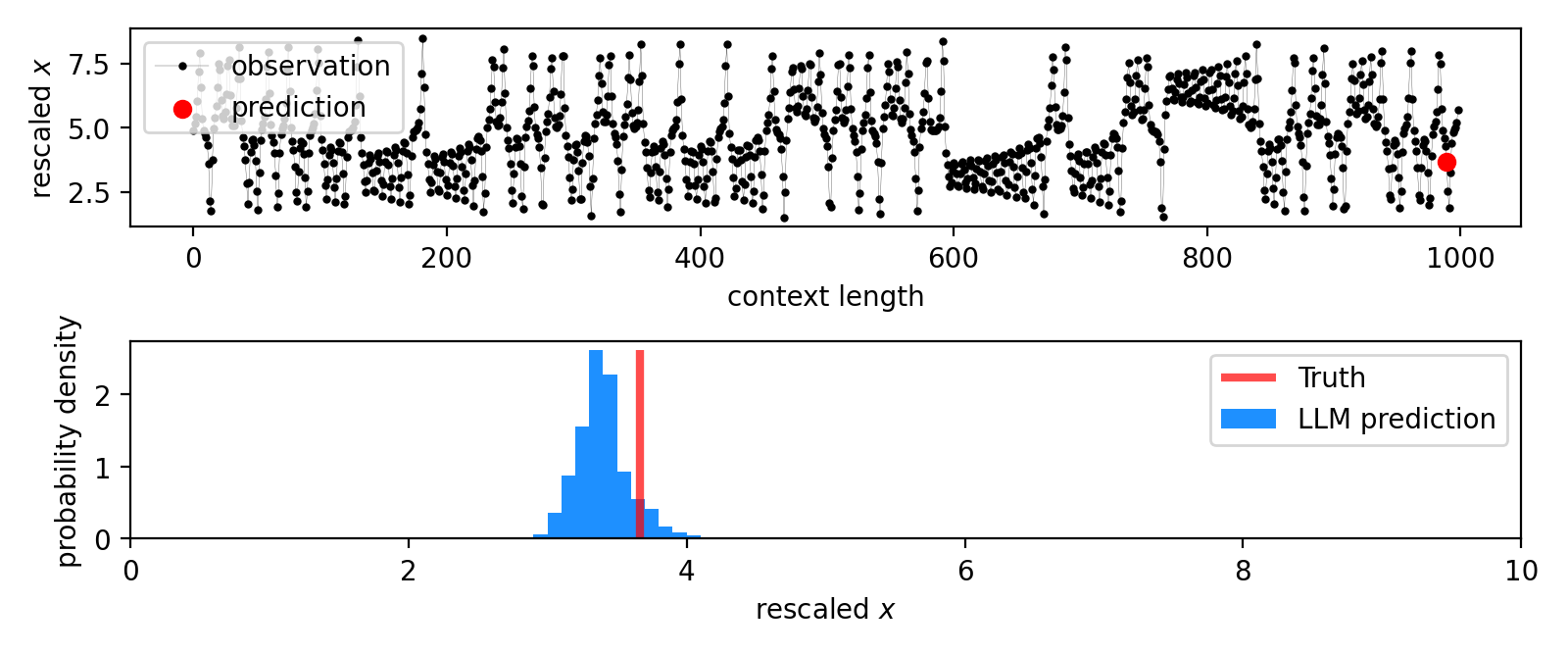}
    \includegraphics[width=\textwidth]{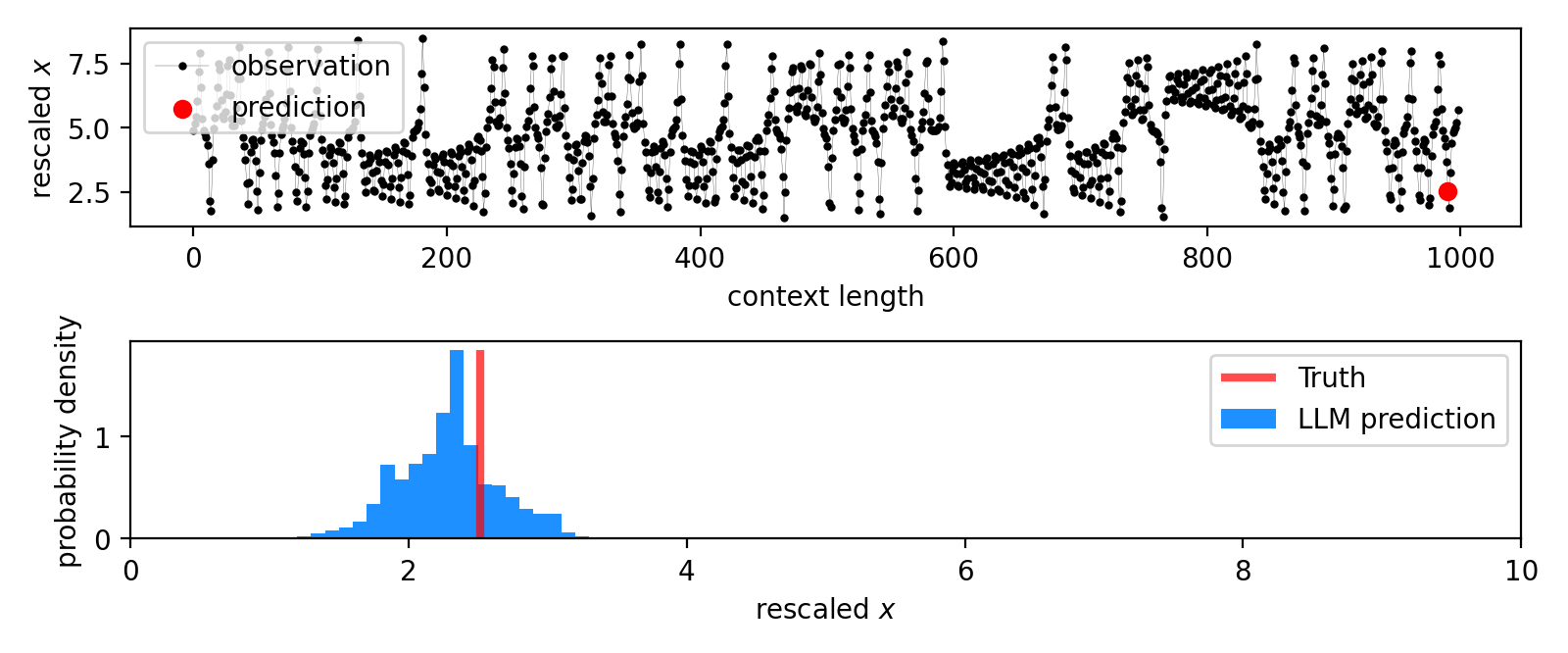}
    \caption{4 consecutive states in a Lorenz system trajectory. Ground truth is shown in red and LLaMA predictions in blue.}
    \label{Lorenz_snapshots}
  \end{minipage}
\end{figure}

One may naively remark upon the possibility that the in-context learning task for the Lorenz system and the logistic map could be rendered trivial if $x_{t+1}$ always falls close to $x_{t}$, in which case the LLM only needs to learn a static noisy distribution in order to decrease the loss. This is not the case with our experiments.
In this section, we demonstrate the non-triviality of the learning tasks in \cref{noisy_logistic_map_snapshots,Lorenz_snapshots}. In both cases, it is clear that the LLM has successfully learned to actively predict the expected mean position of the next state, and, in the logistic map example, the variance of the next state distribution as well. We note that the Lorenz system is simulated deterministically, hence the true next-state distribution is represented as a delta-function.

\subsection{In-context neural scaling law} \label{neural_scaling}

\begin{figure}[htbp]
  \centering
  \includegraphics[width=0.49\textwidth]{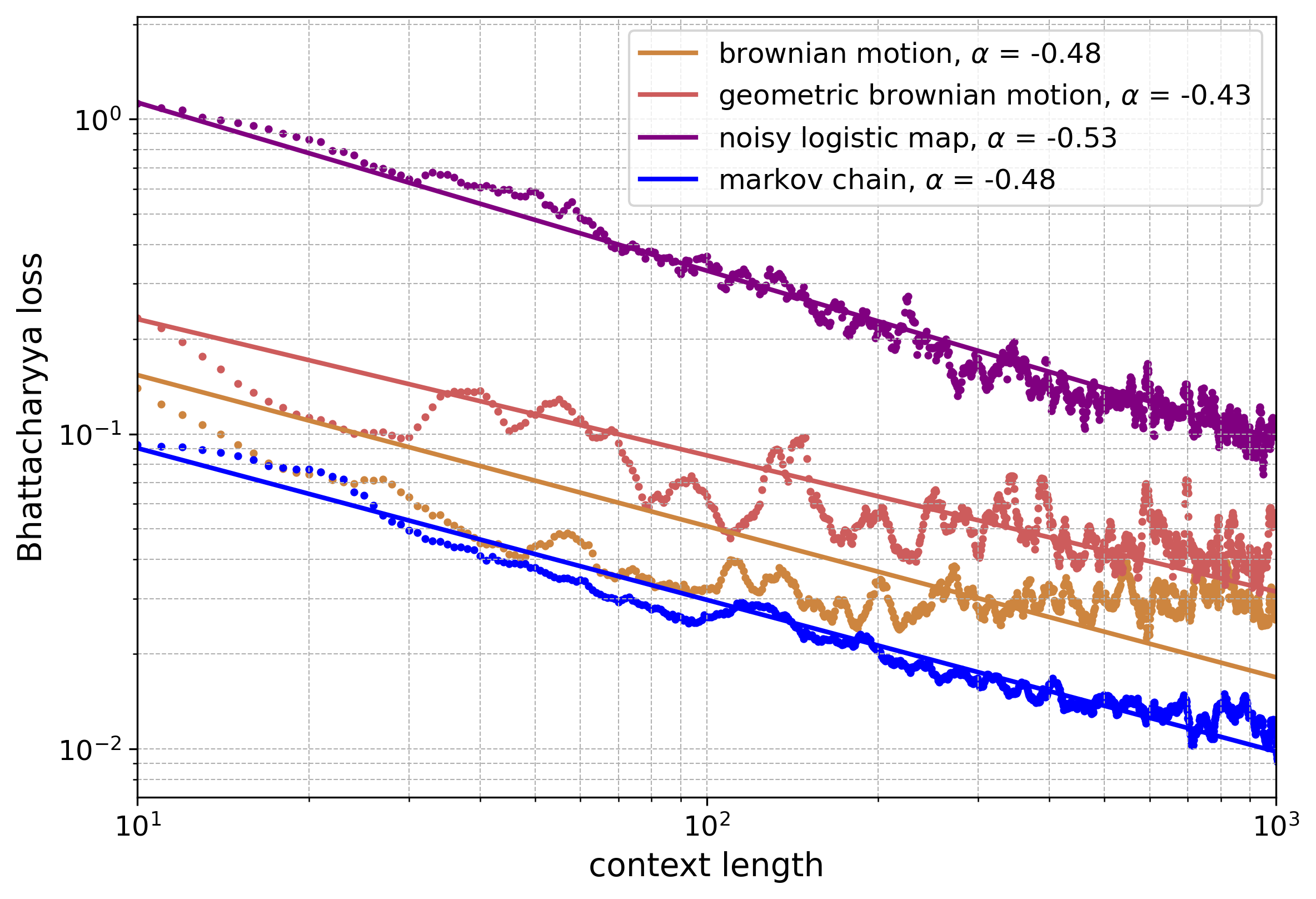}
  \includegraphics[width=0.49\textwidth]{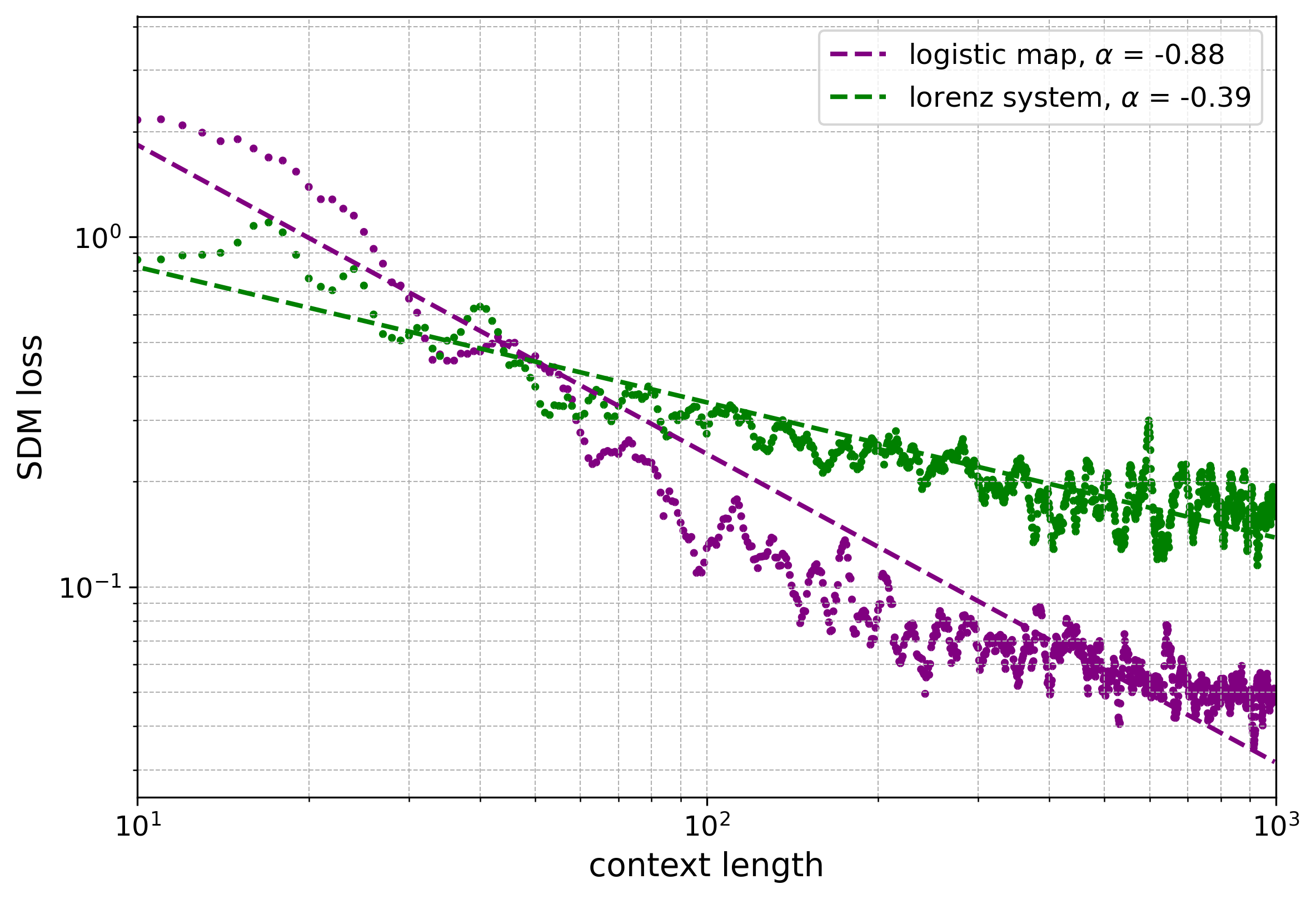}
  \caption{In-context loss curves from LLaMA-13b fitted with power law, with fitted power coefficient $\alpha$ shown in legend. Left: loss of stochastic series measured in Bhattacharyya distance. Right: loss of deterministic series measured in SDM.}
  \label{fig:ICL power law}
\end{figure}

Neural scaling laws~\cite{kaplan2020scaling} describe how the performance of trained neural networks, particularly language models, scales with changes in key factors such as model size ($N$), dataset size ($D$), and computational resources used for training (C). These laws are often observed as power-law relationships in the following form:
\[
  L(N) = \left( \frac{N}{N_{\text{c}}} \right)^{\alpha_N},\quad
  L(D) = \left( \frac{D}{D_{\text{c}}} \right)^{\alpha_D},\quad
  L(C) = \left( \frac{D}{C_{\text{c}}} \right)^{\alpha_C},
\]
where $L$ represents the loss or performance metric of the model. The characteristic factors ($N_{\text{c}}$, $D_{\text{c}}$, and $C_{\text{c}}$), and power coefficient ($\alpha$) are extracted empirically from training curves. The fitted quantities depend on the distribution of data, the model architecture, and the type of optimizer used for training. Such power-law relations appear in log-log plots as straight lines, whose slopes correspond to the parameter $\alpha$. Our loss curves from learning dynamical systems (see \cref{fig:ICL}) reveal an additional neural scaling law that applies to in-context learning:
\[
  L(D_{\text{in}}) = \left( \frac{D_{\text{in}}}{D_{\text{c}}} \right)^{\alpha} ,
\]
where $D_{\text{in}}$ stands for the length of time series observed in the prompt (in-context). In \cref{fig:ICL power law}, we display the fitted power laws to the in-context loss curves.

\subsection{Baselines for noisy logistic map and Markov chains} \label{baseline_models}

In this section, we compare LLM's predictions against baseline models of known architectures, 
in order to understand the difficulty of the in-context learning task and make better sense of the Bhattacharyya loss. 
Specifically, we consider the following baseline models: unigram and bi-gram models for discrete Markov chains, and linear and non-linear autoregressive models with 1-step memory (AR1) for noisy logistic maps. 
The bi-gram model for the Markov chain has an unfair advantage since it is designed to model Markovian processes where the probability distribution of a token depends only on the previous token, i.e., inferring the values of the transition matrix. The unigram model, on the other hand, models all tokens as drawn i.i.d. from the same distribution.

\begin{figure}[htbp]
  \centering
  \includegraphics[width=0.49\textwidth]
  {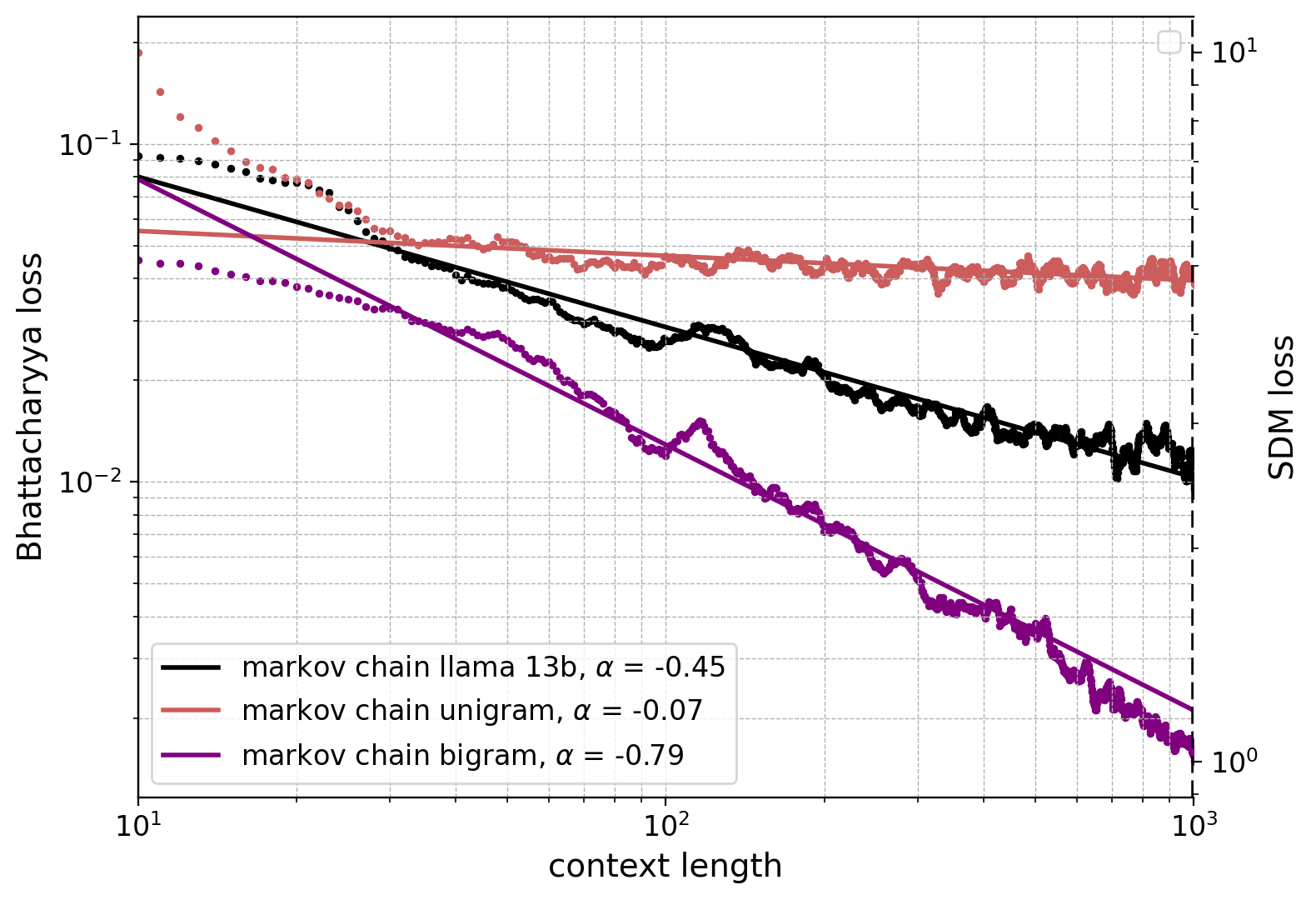}
  \includegraphics[width=0.49\textwidth]
  {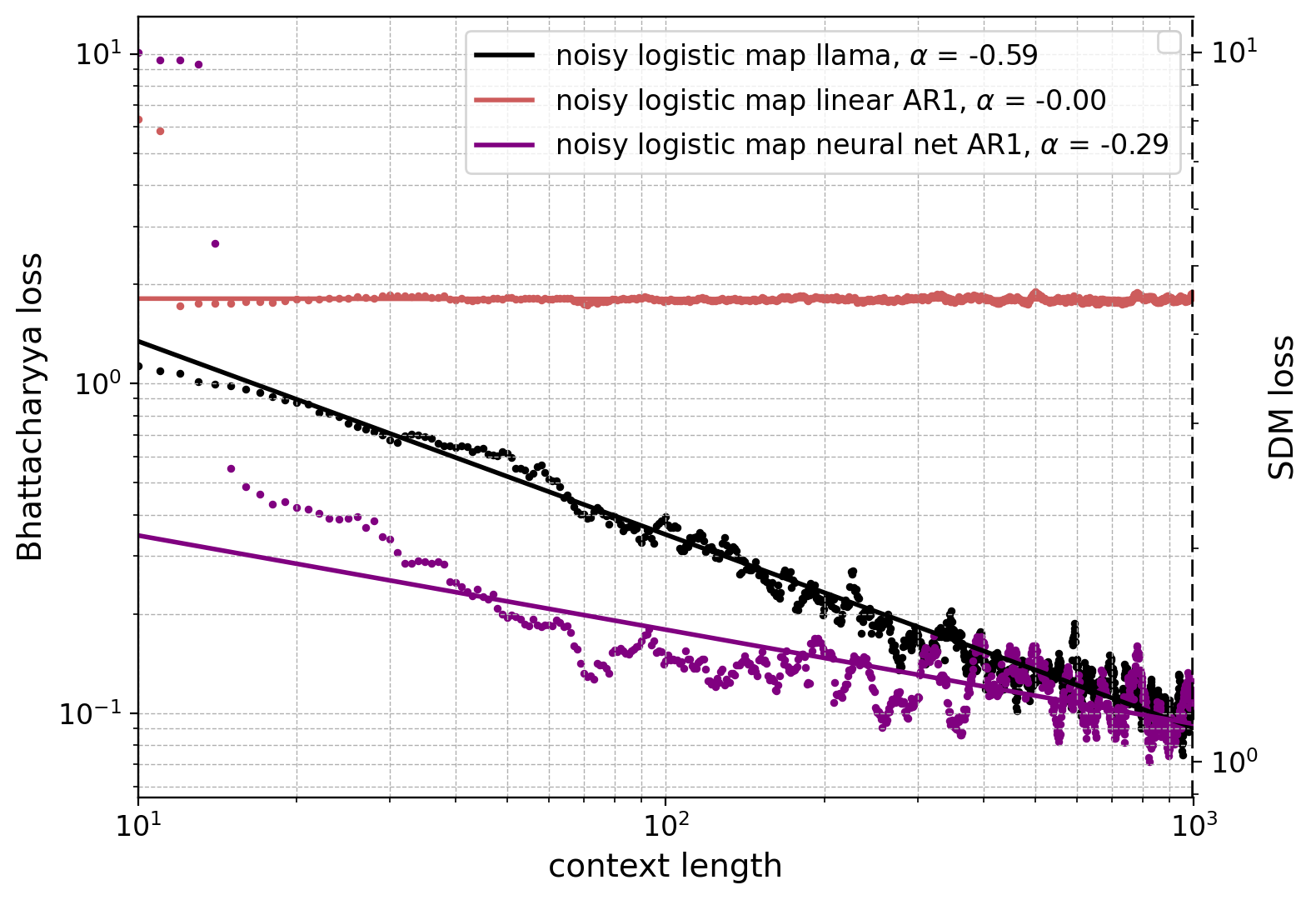}
  \caption{LLM in-context loss curves against the baseline model loss curves. The coefficient $\alpha$ denotes the fitted scaling exponent.}
  \label{baseline_plots}
\end{figure}

The neural network AR1 model takes a state $x_{t-1}$ as input, and outputs prediction for next state $x_t$ as a Gaussian distribution parameterized by mean and variance: $f_\theta: x_{t} \rightarrow \mathcal{N}(\mu_\theta(x_t), \sigma_\theta(x_t))$. As such, it also has the unfair advantage of hard-coded Gaussianiety. LLaMA, on the other hand, must infer the correct distribution family from data. Despite this intrinsic disadvantage, LLaMA still outperforms the neural network AR1 model in the large context limit. The NN used in the non-linear AR1 models features three fully connected hidden layers of widths 64, 32, and 16. We found that simpler neural networks are easily trapped in local minima, leading to unstable performance. The loss curves in \cref{baseline_plots} are obtained by training an independent copy of this neural network to convergence at each context length, and predict the next state distribution using the trained NN. The training loss is defined as the negative log-likelihood of the observation data:
\begin{align*}
  \nonumber \mathcal{L}(\textup{data}, \theta) & = -\sum_{{x_{t-1}, x_t}\in \textup{data}} \log P ( x_t; \mu_\theta(x_{t-1}), \sigma_\theta(x_{t-1}) )                                                         \\
                                               & = \sum_{{x_{t-1}, x_t}\in \textup{data}} \log \sigma_\theta(x_{t-1}) - \frac{1}{2} \left( \frac{x_t - \mu_\theta(x_{t-1})}{\sigma_\theta(x_{t-1})} \right)^2.
\end{align*}

\begin{figure}[htbp]
  \centering
  \includegraphics[width=0.32\textwidth]
  {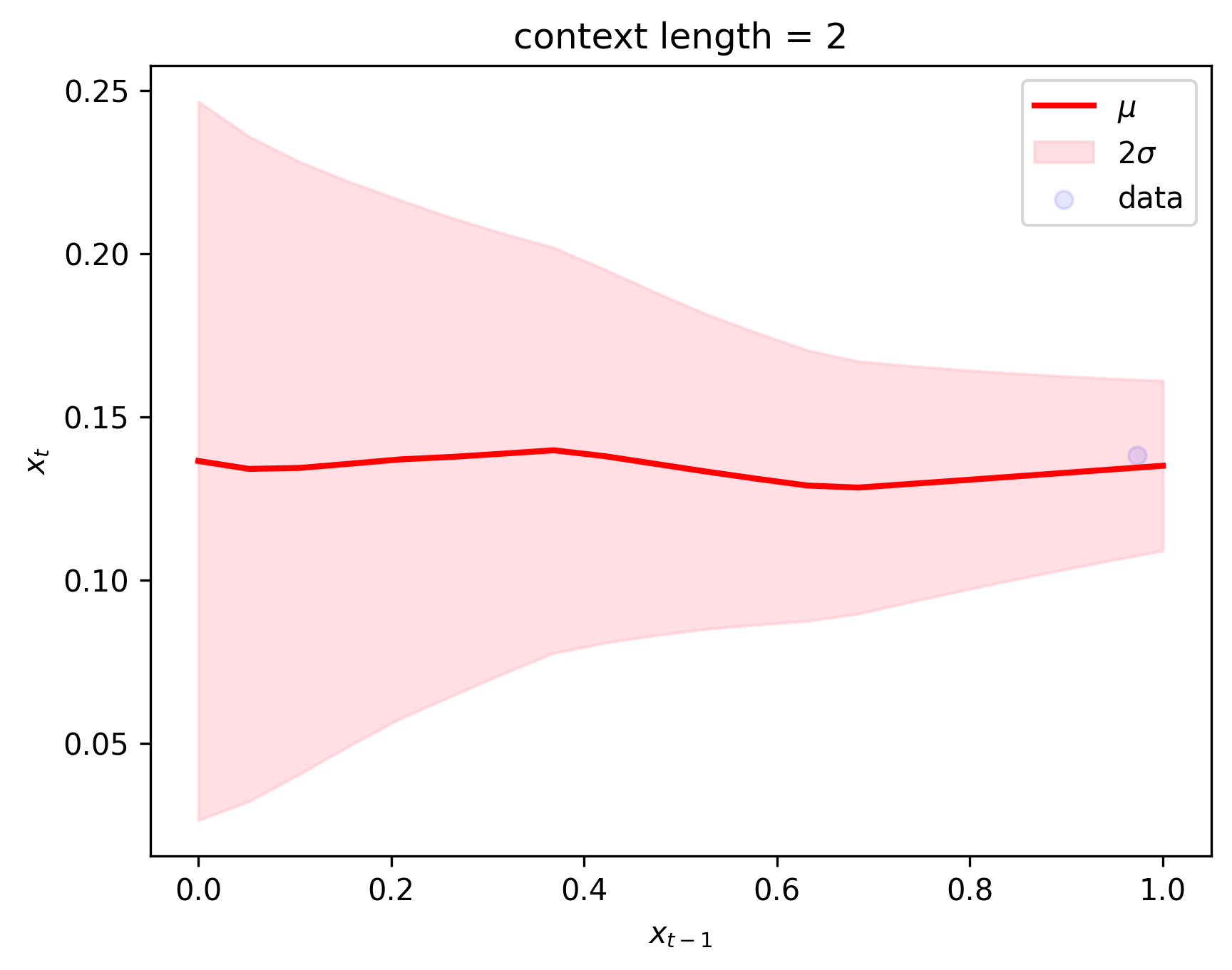}
  \includegraphics[width=0.32\textwidth]
  {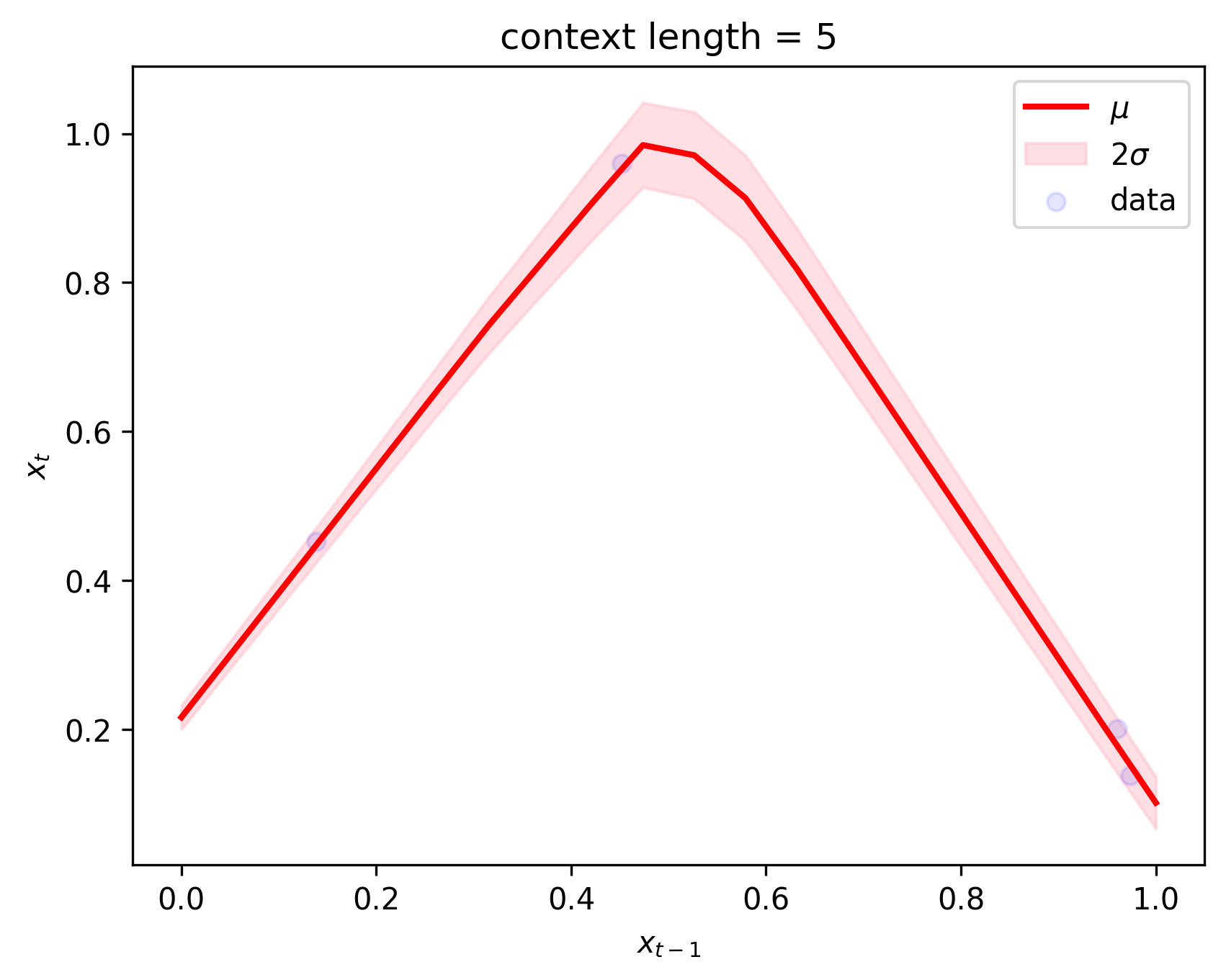}
  \includegraphics[width=0.32\textwidth]
  {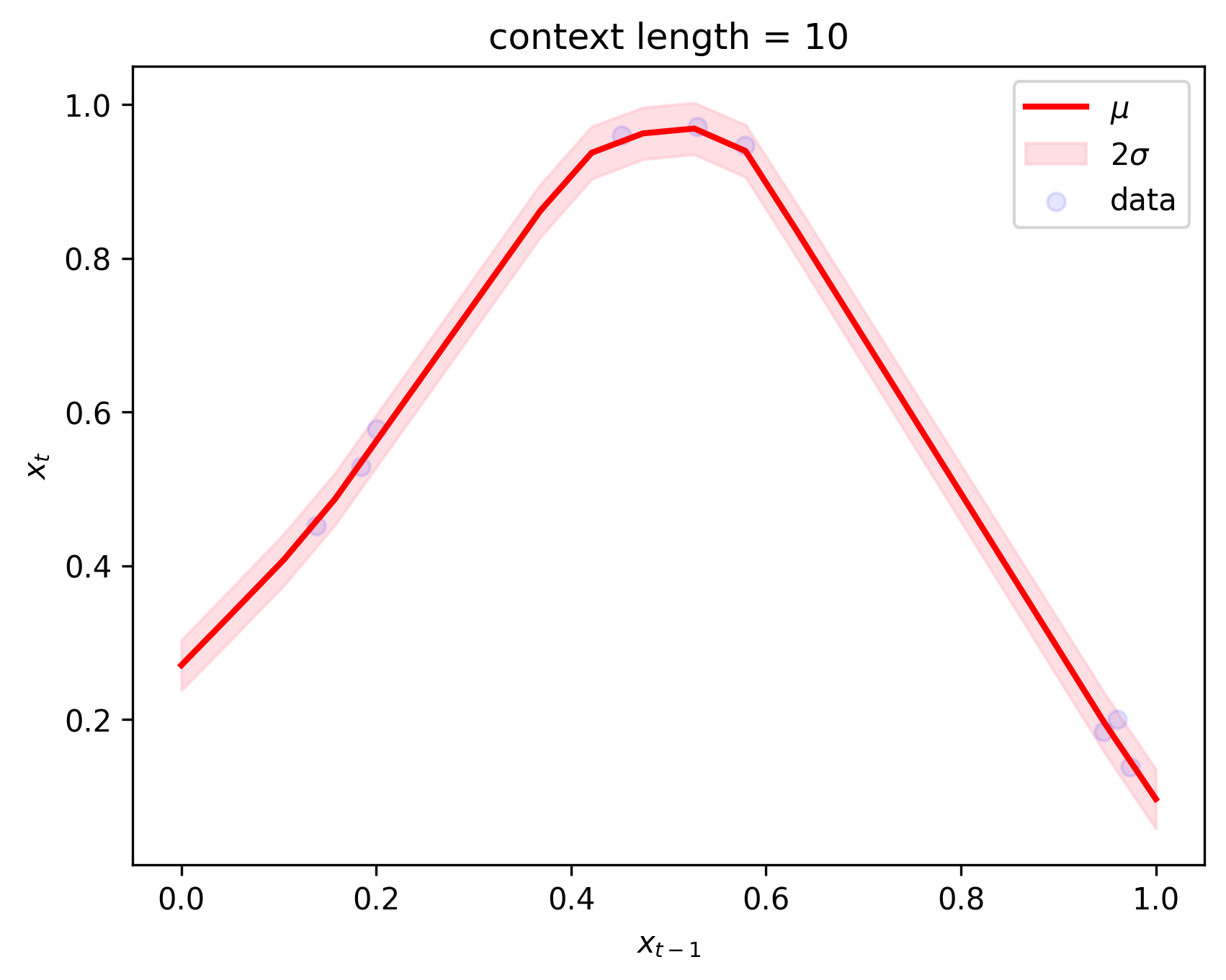}
  \\
  \includegraphics[width=0.32\textwidth]
  {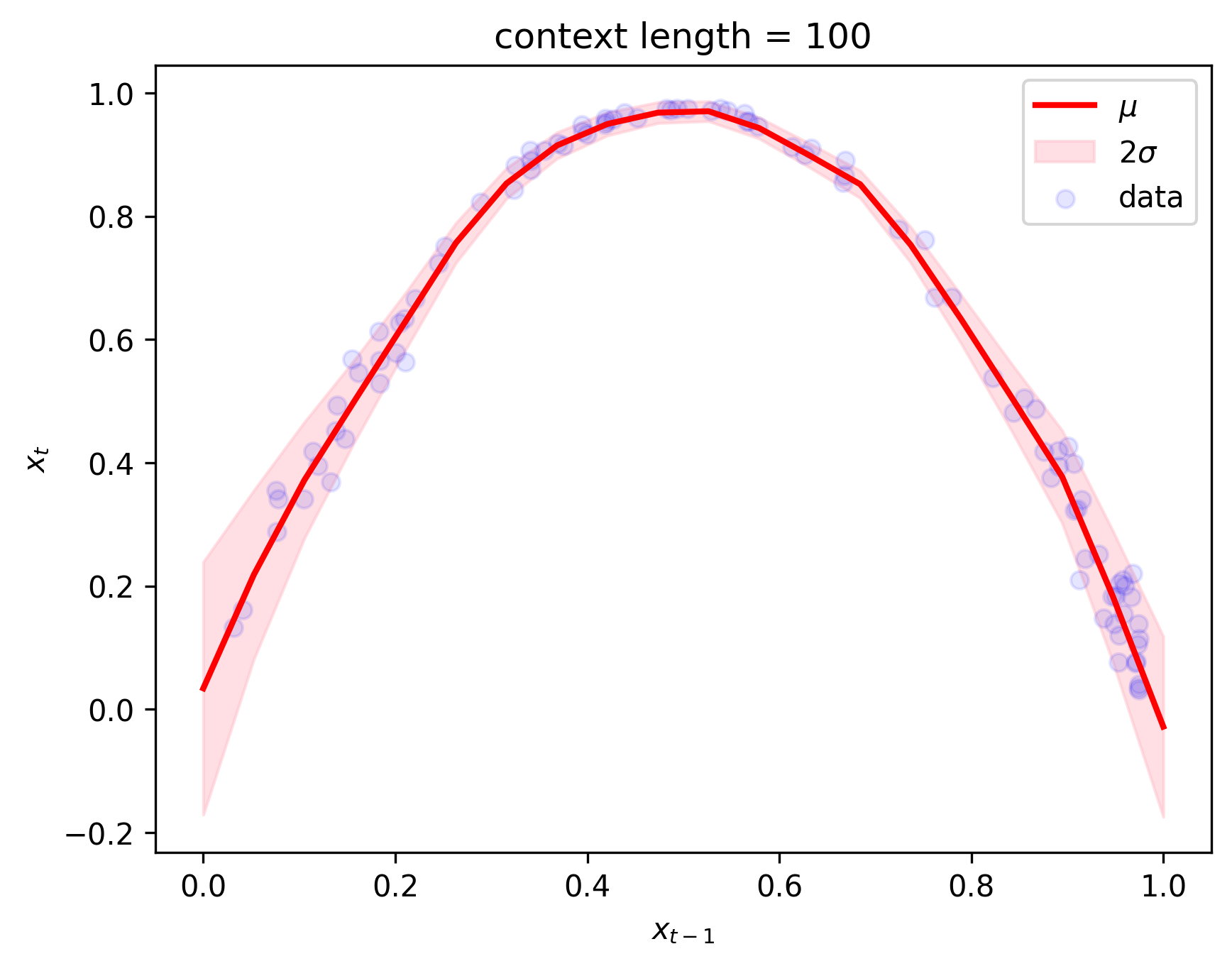}
  \includegraphics[width=0.32\textwidth]
  {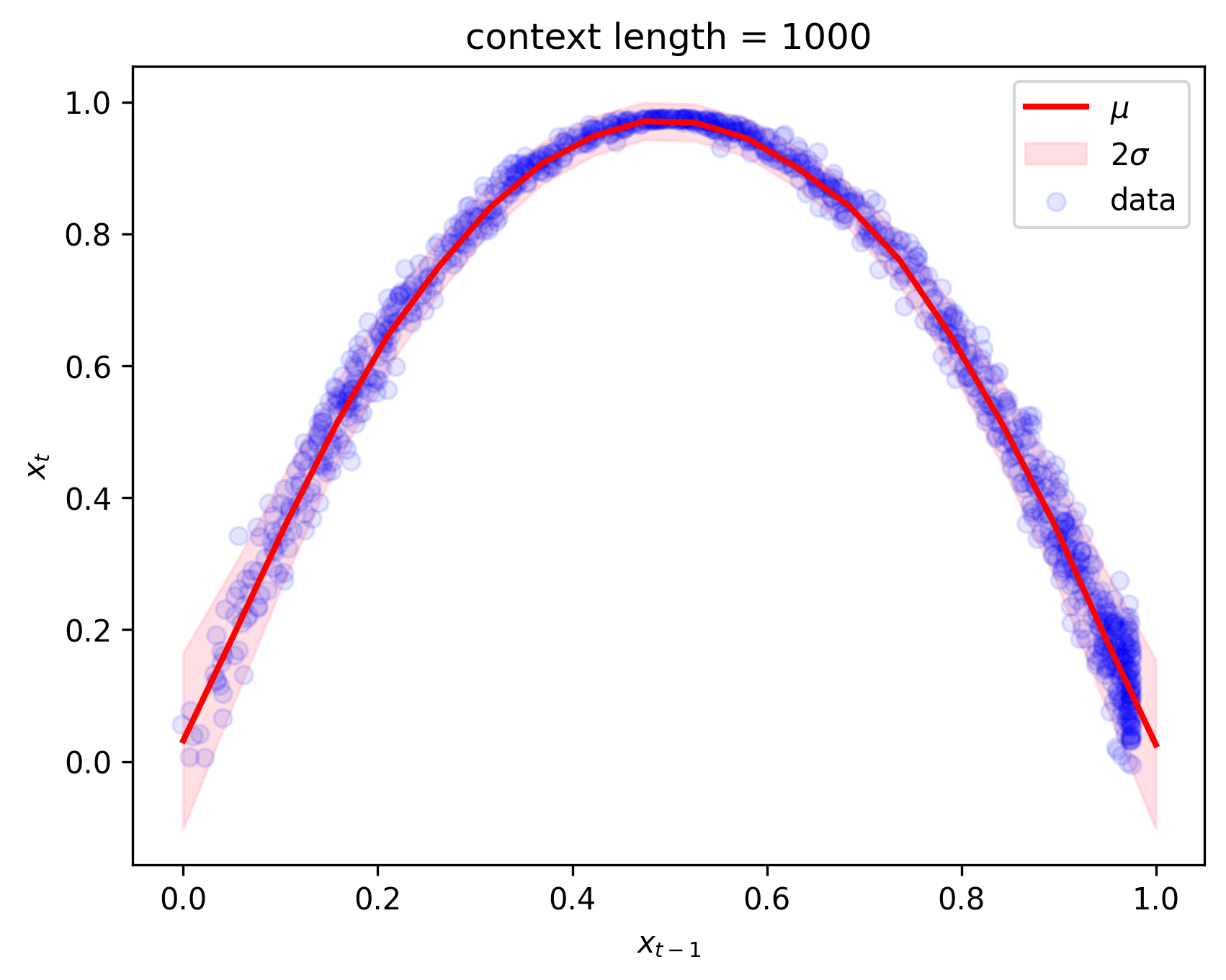}
  \includegraphics[width=0.32\textwidth]
  {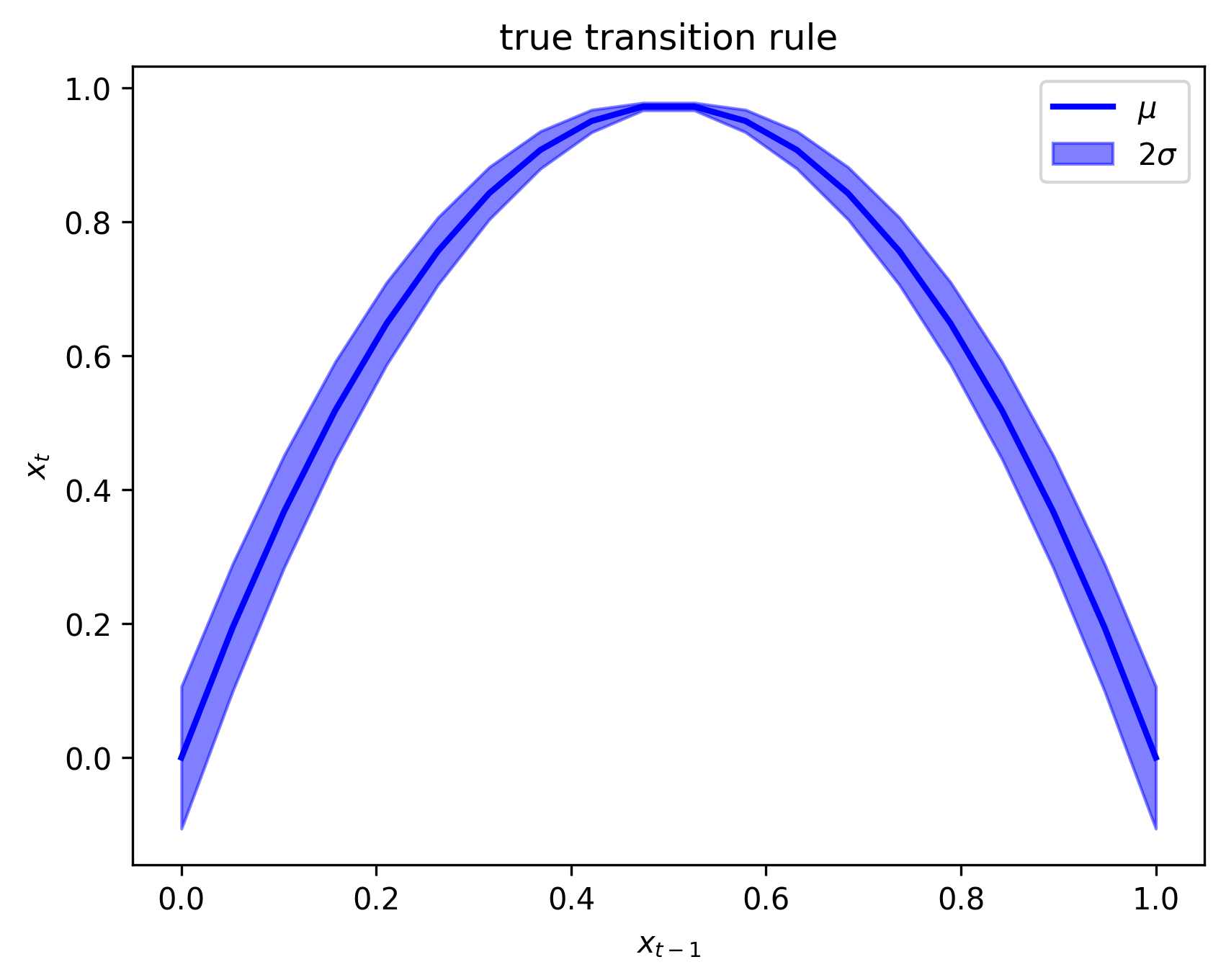}
  \caption{Noisy logistic map transition rules, $P(x_t|x_t-1)$, learned by a neural network-based AR1 model against the ground truth transition rule.}
  \label{NN_transition_rules}
\end{figure}

Furthermore, the ensemble of NNs allows us to visualize the learned transition functions $P(x_{t+1}|x_{t})$ at each context length. In \cref{NN_transition_rules}, we show how the transition rules learned by the NNs gradually converge to the ground truth as context length increases. Since at large context length, the LLMs achieve similar loss as the NN-based AR1 model, it is reasonable to expect the LLM to have learned a transition function of similar accuracy as shown in the 5th plot in \cref{NN_transition_rules}. However, it is difficult to visualize the full transition rules $P(x_{t+1}|x_{t})$, for $x_{t} \in [0,1]$, as learned by an LLM, because doing so would require appending an array of $x_t$s at the end of a training sequence, which would render the training sequence incorrect.

\subsection{Invariant measure and the early plateauing of in-context loss} \label{app: invariant measure}
While most datasets are well-described by the power laws, two loss curves --- the Brownian motion and geometric Brownian motion --- plateau early at a context length of about $10^2$, as shown in \cref{fig:ICL power law}. We attribute this early plateauing to the fact that the Brownian and geometric Brownian motions ``wander out of distribution" at large time $t$, while all other dynamical systems studied in this paper converge to stable distributions (i.e., the invariant measure). A Markovian system (stochastic or deterministic) governed by a transition rule $P(x_{t+1}|x_t)$ is said to have an invariant measure $\pi$ if
\begin{equation}\label{invariant measure for continuous markovian system}
  \pi(x_{t+1}) = \int_{\mathcal{X}} \pi(x_t) P(x_{t+1}|x_t) \,\textup{d}x_t, \quad x_{t+1} \in \mathcal{X}.
\end{equation}
If a system is initialized by $\pi(x)$ and evolves according to $P$, then the distribution of states at the next step will still follow $\pi(x)$. This property makes $\pi$ an invariant or stationary distribution for the system. It has been shown that the logistic map and Lorenz systems in the chaotic regime converge almost surely to their respective invariant measures, regardless of the initialization~\cite{strogatz2018nonlinear}.

For discrete Markov chains governed by a transition matrix, $p$, the stationary distribution is defined as a discrete probability mass function, denoted by $\vec{\pi}$, such that
\begin{equation}\label{invariant measure for discrete markovian syste}
  \vec{\pi}=p\vec{\pi},
\end{equation}
which is analogous to the continuous case described by \cref{invariant measure for continuous markovian system}. By definition, any non-negative right eigenvector of $p$ with eigenvalue $\lambda=1$ is a stationary distribution of $p$. \cite{Sethna2021} showed that a valid transition matrix has at least one stationary distribution. On the other hand, neither the Brownian nor the geometric Brownian motion has invariant distributions\footnote{For stochastic systems, the invariant measure is sometimes referred to as the stationary distribution.} on unbounded domains (e.g., when $\mathcal{X}=\mathbb{R}$). This can be seen from the marginalized distribution $P(x_t)$ at time $t$. For the Brownian motion defined in \cref{eq_brownian}, the marginalized distribution of $x_t$ at time $t$ is a normal distribution:
\begin{align} \label{marginalized PDF BM}
  P(x_t)=\frac{1}{\sqrt{2\pi \sigma^2 t}}\exp\left(-\frac{(x_t-\mu t)^2}{2 \sigma^2 t} \right),
\end{align}
while for the geometric Brownian motion defined in \cref{GBM_SDE}, the marginalized distribution of $x_t$ is a log-normal distribution~\cite{crow1987lognormal}:
\begin{align} \label{marginalized PDF GBM}
  P(x_t)=\frac{1}{x_t\sqrt{2\pi \sigma^2 t}}\exp\left(-\frac{(\log(\frac{x_t}{x_0})-\mu t - \frac{\sigma^2}{2}t)^2}{2 \sigma^2 t} \right).
\end{align}
Both \cref{marginalized PDF BM,marginalized PDF GBM} are time-dependent and do not converge to a stationary distribution in the limit $t \rightarrow \infty$. For the Brownian and geometric Brownian motions, the LLM might decide to only consider the most recent segment of time steps, and ignore the earlier data, which are in some sense ``out of distribution''. This could explain the early plateauing of loss curves. Indeed, the classical neural scaling laws can be improved or broken if the scheduling of the training data shifts in distribution, as shown in \cite{sorscher2023neural}. Different from \cite{sorscher2023neural,lu2023grab}, which alter the scheduling of data to achieve better learning curves that decrease faster with the size of training data, our experiments consider time series with pre-determined transition laws. Therefore, we cannot tamper with the scheduling of our data to make it stationary without altering the transition rules underlying the time series.

\subsection{Temperature and variance} \label{app:temperature}
\begin{figure}[htbp]
  \begin{minipage}[b]{0.47\linewidth}
    \centering
    \includegraphics[width=\textwidth]{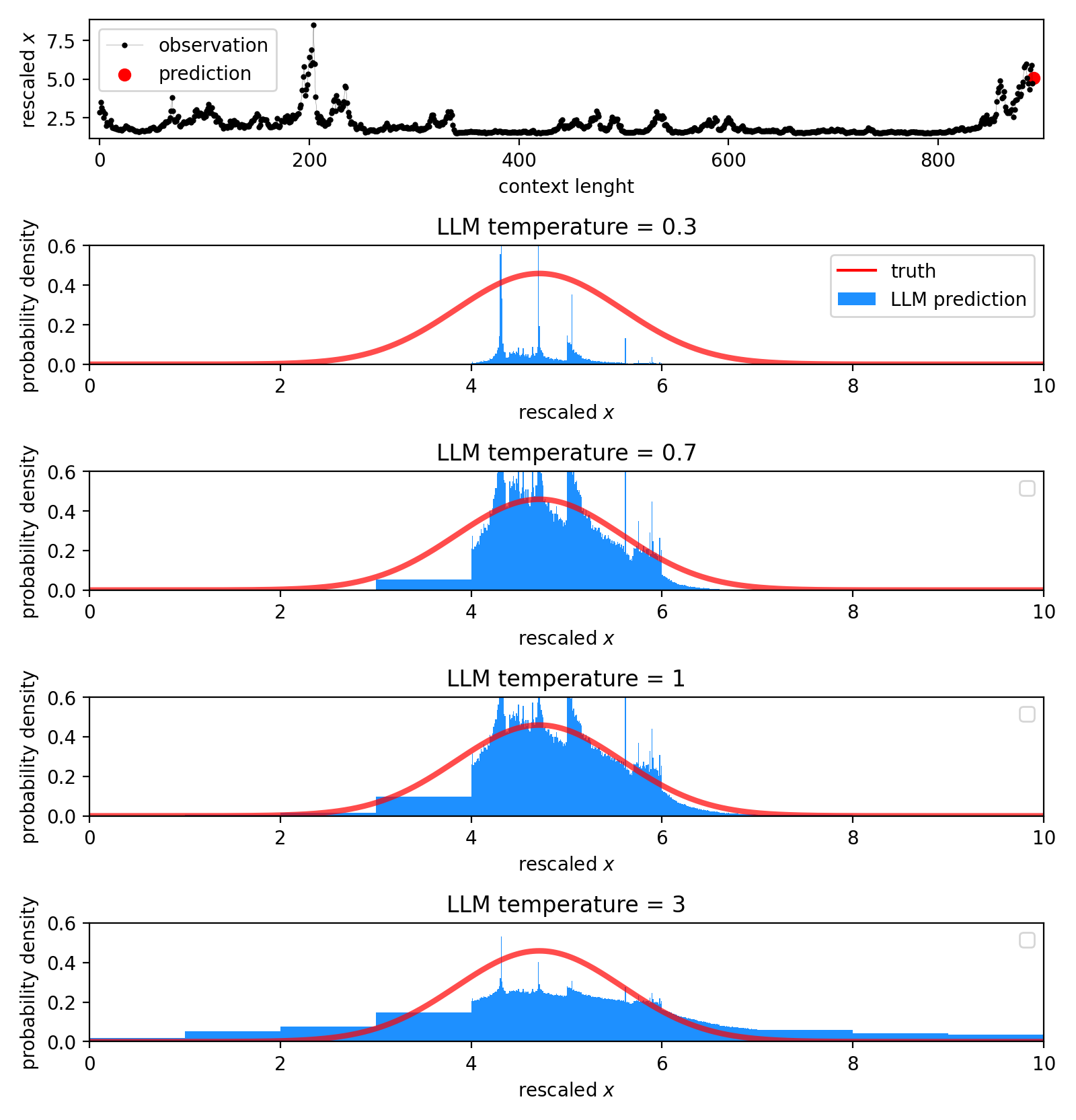}
    \vspace{-2em}
    \caption{Next state prediction for Geometric Brownian motion.
      Topmost: Input stochastic time series (black), and the state to be predicted (red).
      Rest: The Hierarchy-PDF prediction extracted from LLaMA-13b evaluated at different temperatures ranging from $T=0.3$ to $3$, along with ground truth PDF (red).
    }
    \label{fig:GBM_example_temperature_scaling}
  \end{minipage}
  \hspace{1em} 
  \begin{minipage}[b]{0.47\linewidth}
    \centering
    \includegraphics[width=\textwidth]{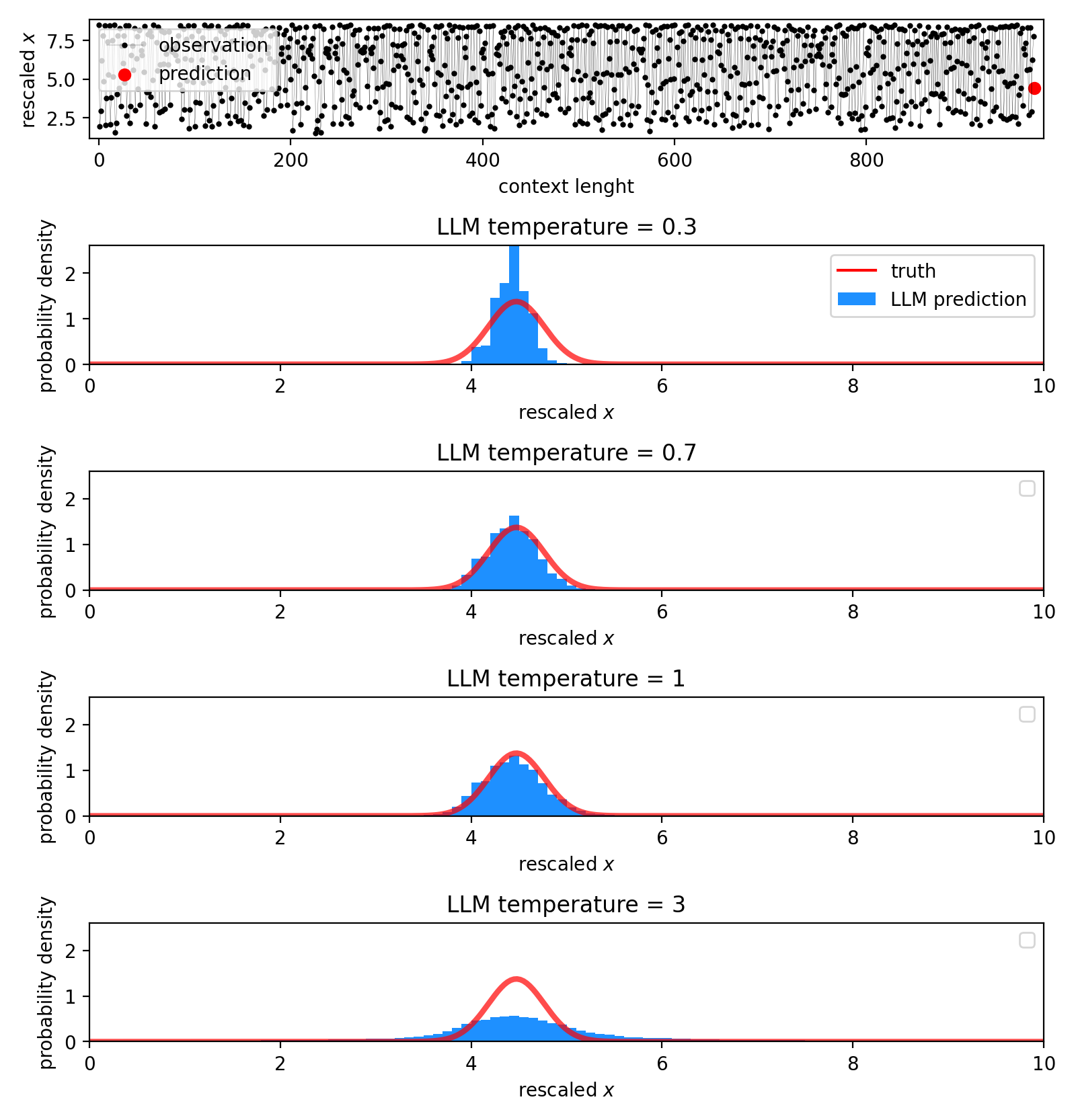}
    \vspace{-2em}
    \caption{Next state prediction for the noisy logistic map.
      Topmost: Input stochastic time series (black), and the state to be predicted (red).
      Rest: The Hierarchy-PDF prediction extracted from LLaMA-13b evaluated at different temperatures ranging from $T=0.3$ to $3$, along with ground truth PDF (red).}
    \label{fig:noisy_logistic_example_temperature_scaling}
  \end{minipage}
\end{figure}

The temperature $T$ is a hyper-parameter that controls the variance of the softmax output layer. Although most LLMs are trained at $T=1$, it is common practice to tune the temperature in the interval $T\in [0.8,1.2]$ during inference. Then, one can opt for increased diversity (high $T$), or better coherence (low $T$) in the generated output. The temperature hyper-parameter affects the uncertainty, or variance, in the Hierarchy-PDF extracted from the LLM. \cref{fig:GBM_example_temperature_scaling,fig:noisy_logistic_example_temperature_scaling} show how different temperatures change the shape of the Hierarchy-PDF. In both cases, higher temperature leads to higher variance in the PDF.

\begin{figure}[htbp]
  \centering
  \includegraphics[width=0.7\textwidth]{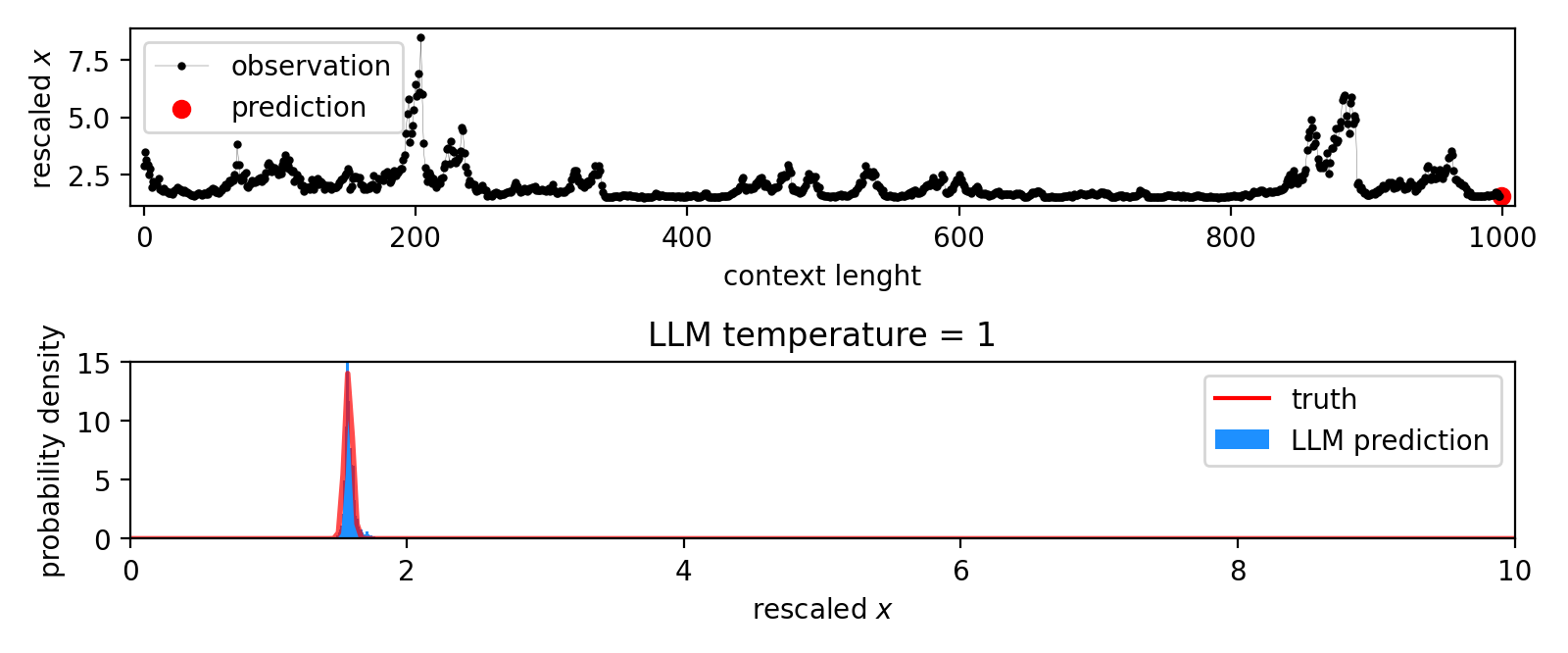}
  \vspace{-1em}
  \caption{Most data points in GBM are trapped in low variance region with small $X_t$. The hierarchy-PDF must be very refined to resolve these minuscule variances.}
  \label{fig:GBM_example_temperature_scaling_state_999}
\end{figure}

We highlight the different refinement schemes used in these figures: for GBM, the PDF is refined to the last (third) digit near the mode, and left coarse elsewhere else. This is because the true variance for GBM can span two orders of magnitude (see \cref{fig:GBM sigma}), with most data points trapped in the low-variance region at small $X_t$. Hence, we require high precision to resolve these small variances in \cref{fig:GBM_example_temperature_scaling_state_999}. On the other hand, the noisy logistic map time series does not suffer from this issue, and thus we uniformly refine its PDF only up to the second digit.

\begin{figure}[htbp]
  \begin{minipage}[b]{0.47\linewidth}
    \centering
    \footnotesize\text{LLM temperature = 0.3}
    \includegraphics[width=\textwidth]{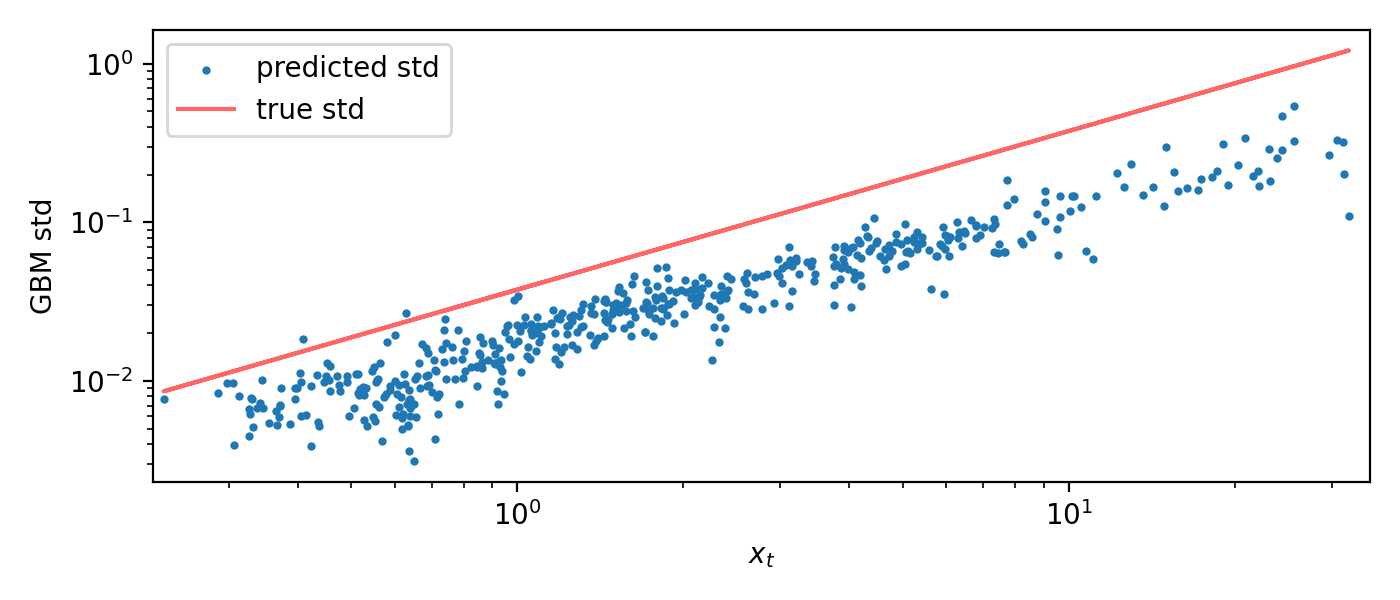}
    \footnotesize\text{LLM temperature = 0.7}
    \includegraphics[width=\textwidth]{figures/GBM_sigma_temp0.7.png}
    \footnotesize\text{LLM temperature = 1}
    \includegraphics[width=\textwidth]{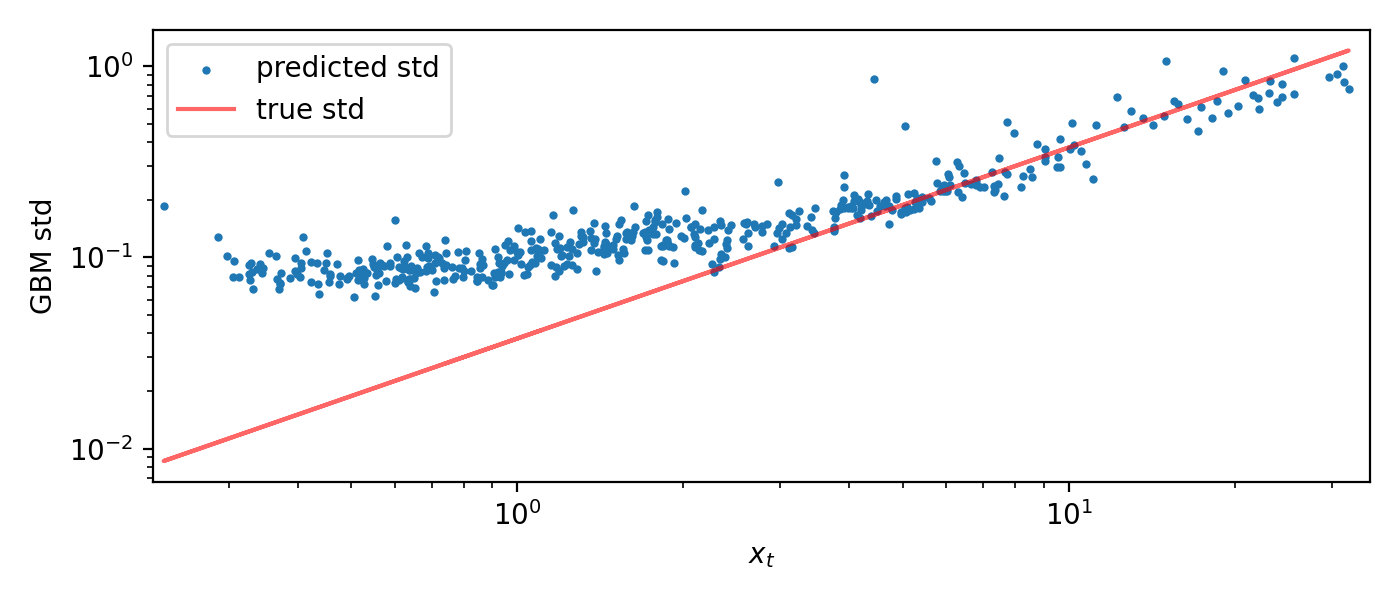}
    \vspace{-1em}
    \caption{GBM standard deviation $\sigma$ as a function of state value $X_t$, learned by the LLM, along with the ground truth. The LLM prediction is evaluated at temperatures ranging from $T=0.3$ to $1$.
    }
    \label{fig:GBM_sigma_temp_scaling}
  \end{minipage}
  \hspace{1em} 
  \begin{minipage}[b]{0.47\linewidth}
    \centering
    \footnotesize\text{LLM temperature = 0.3}
    \includegraphics[width=\textwidth]{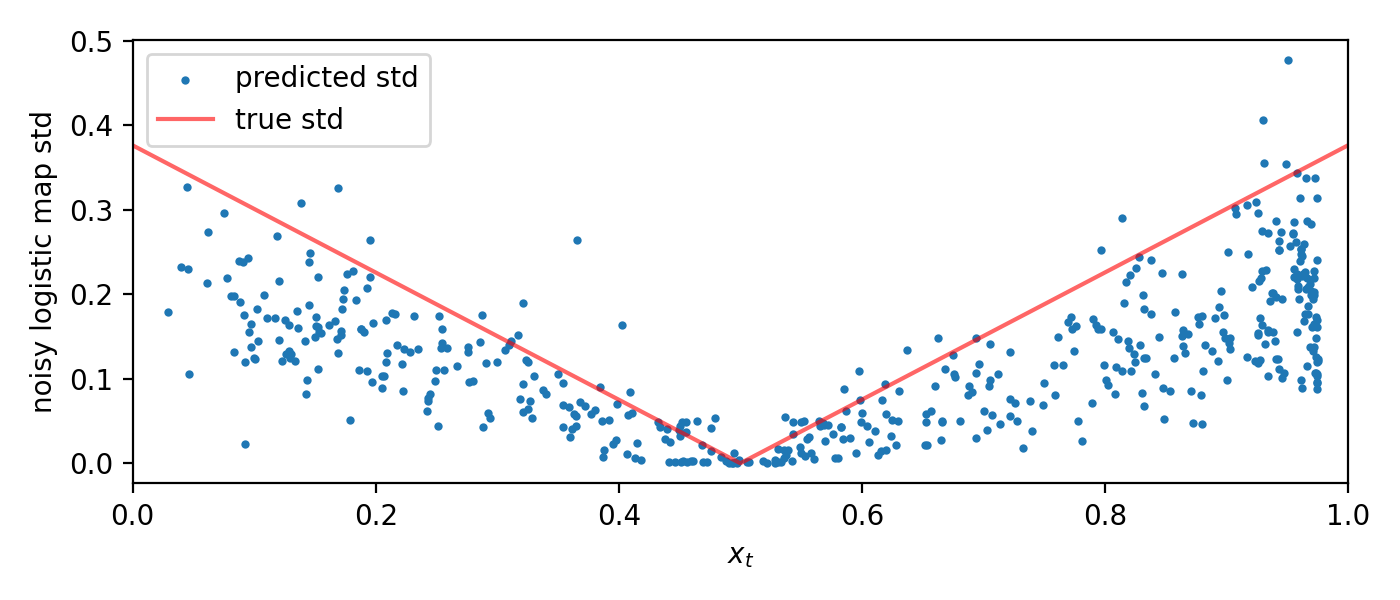}
    \footnotesize\text{LLM temperature = 0.7}
    \includegraphics[width=\textwidth]{figures/noisy_logistic_sigma_temp0.7.png}
    \footnotesize\text{LLM temperature = 1}
    \includegraphics[width=\textwidth]{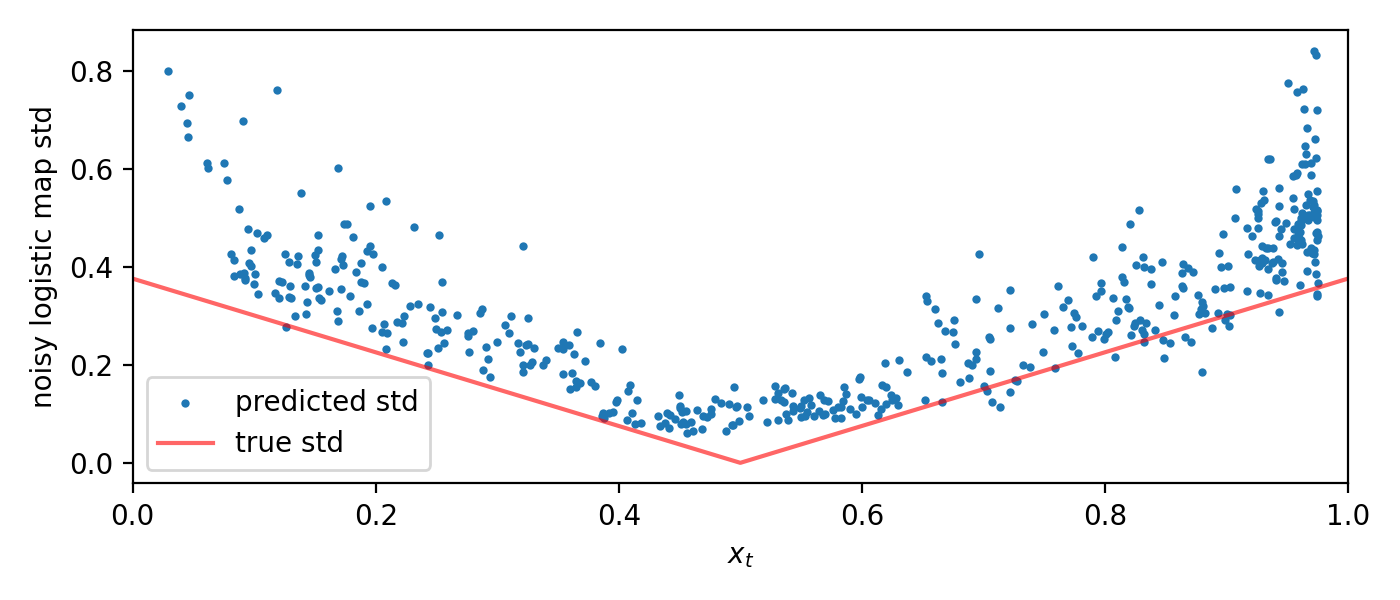}
    \vspace{-1em}
    \caption{Noisy logistic map standard deviation $\sigma$ as a function of state value $X_t$, learned by the LLM, along with the ground truth. The LLM prediction is evaluated at temperatures ranging from $T=0.3$ to $1$.
    }
    \label{fig:noisy_logistic_sigma_temp_scaling}
  \end{minipage}
\end{figure}

While the loss curves in our paper are calculated at $T=1$, the predicted $\sigma$ shown in \cref{fig:GBM sigma,fig:logistic sigma} are extracted at $T=0.7$. We performed a grid search on the temperature ranging from $T=0.3$ to $T=3$ (see \cref{fig:GBM_sigma_temp_scaling,fig:noisy_logistic_sigma_temp_scaling}), and observed that $T=0.7$ consistently results in better prediction quality of the variance.

\newpage
\subsection{Hierarchy PDF}
This section documents all three parts of the Hierarchy-PDF algorithm. We refer to the GitHub repository for further details.

\label{app:hierarchy PDF algorithm}

\begin{algorithm}[htbp]
  \caption{Refine Each State in a Stochastic Sequence}
  \begin{algorithmic}
    \STATE {\bfseries Input:}
    \STATE - $S_{\text{traj}}$: A string representing a sampled stochastic trajectory whose states are separated by commas.
    \STATE - $L_{\text{PDF}}$: List of unrefined PDFs for each state.
    \STATE - $KV_{\text{cache}}$: Key-value cache of running $\texttt{model.forward}(S_{\text{traj}})$.
    \FOR{each \texttt{state} and \texttt{PDF} in $S_{\text{traj}}$ and $L_{\text{PDF}}$}
    \STATE $\texttt{PDF} \gets $ RecursiveRefiner(True, \texttt{state}, $D_c$, $D_t$, $KV_{\text{cache}}$)
    \ENDFOR
  \end{algorithmic}
\end{algorithm}

\begin{algorithm}[htbp]
  \caption{Detailed Hierarchy-PDF Recursive Refiner}
  \begin{algorithmic}
    \STATE {\bfseries Input:} Object \texttt{multi\_PDF} representing unrefined PDF using bins of various widths
    \STATE {\bfseries Procedure:} RecursiveRefiner(mainBranch, sequence, $D_c$, $D_t$, $KV_{\text{cache}}$)
    \begin{ALC@g}
      \IF{$D_c = D_t$}
      \STATE \textbf{return} \COMMENT{Terminate if target refinement depth is reached}
      \ENDIF
      \IF{mainBranch is True}
      \STATE \COMMENT{Launch 9 recursive branches if the current sequence is refined}
      \STATE $L_{\text{new}} \gets$ Form 9 new sequences by changing the last digits
      \FOR{each \texttt{sequence} in $L_{\text{new}}$}
      \STATE RecursiveRefiner(False, $\texttt{sequence}, D_c, D_t, KV_{\text{cache}}$)
      \ENDFOR
      \ELSE
      \STATE \COMMENT{Collect refined logits}
      \STATE \texttt{newLogits}, \texttt{newKVcache} $\gets$
      NextTokenProbs($\texttt{sequence}, KV_{\text{cache}}$)
      \STATE Refine \texttt{multi\_PDF} using \texttt{newLogits}
      \ENDIF
      \IF{$D_c + 1 < D_t$}
      \STATE \COMMENT{Launch 10 more branches if $D_t$ not met}
      \STATE $L_{\text{new}} \gets$ Form 10 new sequences by appending digits
      \FOR{each $\texttt{sequence}$ in $L_{\text{new}}$}
      \STATE RecursiveRefiner(False, $\texttt{sequence}, D_c+1, D_t, \texttt{newKVcache}$)
      \ENDFOR
      \ENDIF
    \end{ALC@g}
  \end{algorithmic}
\end{algorithm}

\begin{algorithm}[htbp]
  \caption{Extract Next Token Probabilities}
  \begin{algorithmic}
    \STATE \textbf{function} NextTokenProbs(\texttt{sequence}, \texttt{KVcache}, \texttt{model})
    \begin{ALC@g}
      \STATE \texttt{NextTokenLogit} $\gets$ \texttt{model.forward}(\texttt{sequence}, \texttt{KVcache})[\text{last}] \COMMENT{Extract distribution of next token}
      \STATE Update \texttt{KVcache}
    \end{ALC@g}
    \STATE \textbf{return} \texttt{NextTokenLogit}, \texttt{KVcache}
  \end{algorithmic}
\end{algorithm}



\end{document}